\documentclass[10pt,journal,compsoc]{IEEEtran}

\usepackage{todonotes}
\usepackage{longtable}
\usepackage{tabularx}
\usepackage[colorlinks]{hyperref}
\usepackage{threeparttable}
\usepackage{url}
\makeatletter
\g@addto@macro{\UrlBreaks}{\UrlOrds}
\makeatother
\usepackage{booktabs}
\usepackage{multirow}
\usepackage{amsmath}
\usepackage{amssymb}
\usepackage{color, colortbl}
\definecolor{Gray}{gray}{0.95}
\usepackage{dblfloatfix}
\usepackage{microtype}

\ifCLASSOPTIONcompsoc
  \usepackage[nocompress]{cite}
\else
  \usepackage{cite}
\fi

\ifCLASSINFOpdf
\else
\fi

\begin{document}
\title{Semantic Object Accuracy for\\Generative Text-to-Image Synthesis}

\author{Tobias~Hinz,
		Stefan~Heinrich,
		and Stefan~Wermter%
		\IEEEcompsocitemizethanks{\IEEEcompsocthanksitem The authors are with the Knowledge Technology Group, University of Hamburg, Germany, Email: \{hinz, heinrich, wermter\}@informatik.uni-hamburg.de.}%
}

\markboth{Journal of \LaTeX\ Class Files,~Vol.~14, No.~8, August~2015}%
{Shell \MakeLowercase{\textit{et al.}}: Bare Demo of IEEEtran.cls for Computer Society Journals}

\IEEEtitleabstractindextext{%
\begin{abstract} %
	Generative adversarial networks conditioned on textual image descriptions are capable of generating realistic-looking images.
	However, current methods still struggle to generate images based on complex image captions from a heterogeneous domain.
	Furthermore, quantitatively evaluating these text-to-image models is challenging, as most evaluation metrics only judge image quality but not the conformity between the image and its caption.
	To address these challenges we introduce a new model that explicitly models individual objects within an image and a new evaluation metric called \textit{Semantic Object Accuracy} (SOA) that specifically evaluates images given an image caption.
	The SOA uses a pre-trained object detector to evaluate if a generated image contains objects that are mentioned in the image caption, e.g. whether an image generated from \textit{``a car driving down the street''} contains a car.
	We perform a user study comparing several text-to-image models and show that our SOA metric ranks the models the same way as humans, whereas other metrics such as the Inception Score do not.
	Our evaluation also shows that models which explicitly model objects outperform models which only model global image characteristics.
\end{abstract}

}

\maketitle

\IEEEdisplaynontitleabstractindextext

\IEEEpeerreviewmaketitle

\IEEEraisesectionheading{\section{Introduction}\label{sec:introduction}}

\IEEEPARstart{G}{enerative} adversarial networks (GANs) \cite{goodfellow2014generative} are capable of generating realistic-looking images that adhere to characteristics described in a textual manner, e.g. an image caption.
    For this, most networks are conditioned on an embedding of the textual description.
	Often, the textual description is used on multiple levels of resolution, e.g. first to obtain a course layout of the image at lower levels and then to improve the details of the image on higher resolutions.
	This approach has led to good results on simple, well-structured data sets containing a specific class of objects (e.g. faces, birds, or flowers) at the image center.
	
	Once images and textual descriptions become more complex, e.g. by containing more than one object and having a large variety in backgrounds and scenery settings, the image quality drops drastically.
	This is likely because, until recently, almost all approaches only condition on an embedding of the complete textual description, without paying attention to individual objects.
	Recent approaches have started to tackle this by either relying on specific scene layouts \cite{johnson2018image} or by explicitly focusing on individual objects \cite{hinz2019generating, li2019object}.
	In this work, we extend this approach by additionally focusing specifically on salient objects within the generated image.
	However, generating complex scenes containing multiple objects from a variety of classes is still a challenging problem.
	
	The most commonly used evaluation metrics for GANs, the Inception Score (IS) \cite{salimans2016improved} and the Fr\'{e}chet Inception Distance (FID) \cite{heusel2017gans}, are not designed to evaluate images that contain multiple objects and depict complex scenes.
	In fact, both of these metrics depend on an image classifier (the Inception-Net), which is pre-trained on ImageNet, a data set whose images almost always contain only a single object at the image center.
	They also do not evaluate the consistency between image description and generated image and, therefore, can not evaluate whether a model generates images that actually depict what is described in the caption.
	Even evaluation metrics specifically designed for text-to-image synthesis evaluation such as the R-precision \cite{xu2017attngan} often fail to evaluate more detailed aspects of an image, such as the quality of individual objects.
	
	As such, our contributions are twofold: first, we introduce a novel GAN architecture called \textit{OP-GAN} that focuses specifically on individual objects while simultaneously generating a background that fits with the overall image description.
	Our approach relies on an object pathway similar to \cite{hinz2019generating}, which iteratively attends to all objects that need to be generated given the current image description.
	In parallel, a global pathway generates the background features which later on get merged with the object features.   
	Second, we introduce an evaluation metric specifically for text-to-image synthesis tasks which we call \textit{Semantic Object Accuracy} (SOA).
	In contrast to most current evaluation metrics, our metric focuses on individual objects and parts of an image and also takes the caption into consideration when evaluating an image.
	Image descriptions often explicitly or implicitly mention what kind of objects are seen in an image, e.g. an image described by the caption \textit{``a person holding a cell phone''} should depict both a person and a cell phone.
	To evaluate this, we sample all image captions from the COCO validation set that explicitly mention one of the 80 main object categories (e.e. ``person'', ``dog'', ``car'', etc.) and use them to generate images.
	We then use a pre-trained object detector \cite{redmon2018yolov3} and check whether it detects the explicitly mentioned objects within the generated images.
	We perform a user study over several current text-to-image models and show that SOA is highly compatible with human evaluation whereas other metrics, such as the Inception Score, are not.	
	
	We evaluate several variations of our proposed model as well as several state-of-the-art approaches that provide pre-trained models.
	Our results show that current architectures are not able to generate images that contain objects of the same quality as the original images.
	While some models already achieve results close to or better than real images on scores such as the IS and R-precision, none of the models comes close to generating images that achieve SOA scores close to the real images.
	However, our results and user study also show that models that attend to individual objects in one way or another tend to perform better than models, which only focus on global image semantics.

\section{Related Work}			
    Modern architectures are able to synthesize realistic, high-resolution images of many domains.
	In order to generate images of high resolution many GAN \cite{goodfellow2014generative} architectures use multiple discriminators at various resolutions \cite{zhang2017stackgan++}.
	Additionally, most GAN architectures use some form of attention for improved image synthesis \cite{xu2017attngan} as well as matching aware discriminators \cite{reed2016generative} which identify whether images correspond to a given textual description.
	
	Originally, most GAN approaches for text-to-image synthesis encoded the textual description into a single vector which was used as a condition in a conditional GAN (cGAN) \cite{reed2016generative, zhang2017stackgan++}.
	However, this faces limitations when the image content becomes more complex as e.g. in the COCO data set \cite{lin2014microsoft}.
	As a result, many approaches now use attention mechanisms to attend to specific words of the sentence \cite{xu2017attngan}, use intermediate representations such as scene layouts \cite{johnson2018image}, condition on additional information such as object bounding boxes \cite{hinz2019generating} or perform interactive image refinement \cite{sharma2018chatpainter}.
	Other approaches generate images directly from semantic layouts without additional textual input \cite{karacan2016learning, park2019semantic}or perform a translation from text to images and back \cite{sah2018semantically,qiao2019mirrorgan}.
	
	\textbf{Direct Text-to-Image Synthesis}\ \ Approaches that do not use intermediate representations such as scene layouts use only the image caption as conditional input.
	\cite{reed2016generative} use a GAN to generate images from captions directly and without any attention mechanism.
	Captions are embedded and used as conditioning vector and they introduce the widely adopted matching aware discriminator.
	The matching aware discriminator is trained to distinguish between real and matching caption-image pairs (``real''), real but mismatching caption-image pairs (``fake''), and matching captions with generated images (``fake'').
	\cite{cha2019adversarial} modify the sampling procedure during training to obtain a curriculum of mismatching caption-image pairs and introduce an auxiliary classifier that specifically predicts the semantic consistency of a given caption-image pair.
	\cite{zhang2017stackgan, zhang2017stackgan++} use multiple generators and discriminators and are one of the first ones to achieve good image quality at resolutions of $256\times 256$ on complex data sets.
	\cite{zhang2018photographic} have a similar architecture as \cite{zhang2017stackgan} with multiple discriminators but only use one generator while \cite{huang2019hierarchically} generate realistic high-resolution images from text with a single discriminator and generator.
	
	\cite{xu2017attngan} extend \cite{zhang2017stackgan++} and are the first ones to introduce an attention mechanism to the text-to-image synthesis task with GANs.
	The attention mechanism attends to specific words in the caption and conditions different image regions on different words to improve the image quality.
	\cite{yin2019semantics} extend this and also consider semantics from the text description during the generation process.
	\cite{zhu2019dm} introduce a dynamic memory part that selects ``bad'' parts of the initial image and tries to refine them based on the most relevant words.
	\cite{li2019controllable} refine the attention module by having spatial and channel-wise word-level attention and introduce a word-level discriminator to provide fine-grained feedback based on individual words and image regions.
	\cite{qiao2019learn} decompose the text-to-image process into three distinct phases by first learning a prior over the text-image space, then sampling from this prior, and lastly using the prior to generate the image.
	
	\textbf{Text-to-Image Synthesis with Layouts}\ \ When using more complex data sets that contain multiple objects per image, generating an image directly becomes difficult.
	Therefore, many approaches use additional information such as bounding boxes for objects or intermediate representations such as scene graphs or scene layouts which can be generated automatically \cite{li2019layoutgan, jyothi2019layoutvae, li2019seq}.
	\cite{reed2016generating} and \cite{reed2016learning} build on \cite{reed2016generative} by additionally conditioning the generator on bounding boxes or keypoints of relevant objects.
	\cite{raj2017compositional} decomposition textual descriptions into basic visual primitives to generate images in a compositional manner.
	\cite{johnson2018image} introduce the concept of generating a scene graph based on a caption.
	This scene graph  is then used to generate an image layout and finally the image.
	Similar to \cite{johnson2018image}, \cite{hong2018inferring} use the caption to infer a scene layout which is used to generate images.
	\cite{liu2019learning} predict convolution kernels conditioned on the semantic layout, making it possible to control the generation process based on semantic information at different locations.
	
	Given a coarse image layout (bounding boxes and object labels) \cite{zhao2018image} generate images by disentangling each object into a specified part (e.g. object label) and unspecified part (appearance).
	\cite{hinz2019generating} generate images conditioned on bounding boxes for the individual foreground objects by introducing an object pathway that generates individual objects.
	\cite{li2019object} update the grid-based attention mechanism \cite{xu2017attngan} by combining attention with scene layouts.
	Additionally, an object discriminator is introduced which focuses on individual objects and provides feedback whether the object is at the right location.
	\cite{huang2019realistic} refine the grid-based attention mechanism between word phrases and specific image regions of various sizes based on an initial set of bounding boxes.
	\cite{sun2019image} introduce a new feature normalization method and fine-grained mask maps to generate visually different images from a given layout.
	\cite{li2019pastegan} generate images from scene graphs and allow the model to crop objects from other images to paste them into the generated image.
	\cite{vo2019visual} generate a visual-relation scene layout based on the caption.
	For this, they introduce a dedicated module which generates bounding boxes for objects at a given caption in order to condition the network during the image generation process.
	
	\textbf{Semantic Image Manipulation}\ \ Finally, there are methods that allow humans to directly describe the image in an iterative process or that allow for direct semantic manipulation of images.
	\cite{sharma2018chatpainter} condition generation process on a dialogue describing the image instead of a single caption.
	\cite{hong2018learning} facilitate semantic image manipulation by allowing users to modify image layouts which are then used to generate images.
	\cite{donghoon2018context} allow users to input object instance masks into an existing image represented by a semantic layout.
	\cite{el2018keep} generate images iteratively from consecutive textual commands, \cite{cheng2018sequential} provide interactive image editing based on a current image and instructions on how to update the image, and \cite{li2019storygan} generate individual images for a sequence of sentences.
	\cite{mittal2019interactive} do interactive image generation but do not use text as direct input but instead update a scene graph from text over the course of the interaction.
	\cite{nam2018text, zhou2019text}, and \cite{yu2019simgan} modify visual attributes of individual objects in an image while leaving text irrelevant parts of the image unchanged.

\section{Approach}
\label{sec:approach}
	\begin{figure*}[!t]
	\centering
	\includegraphics[width=\textwidth]{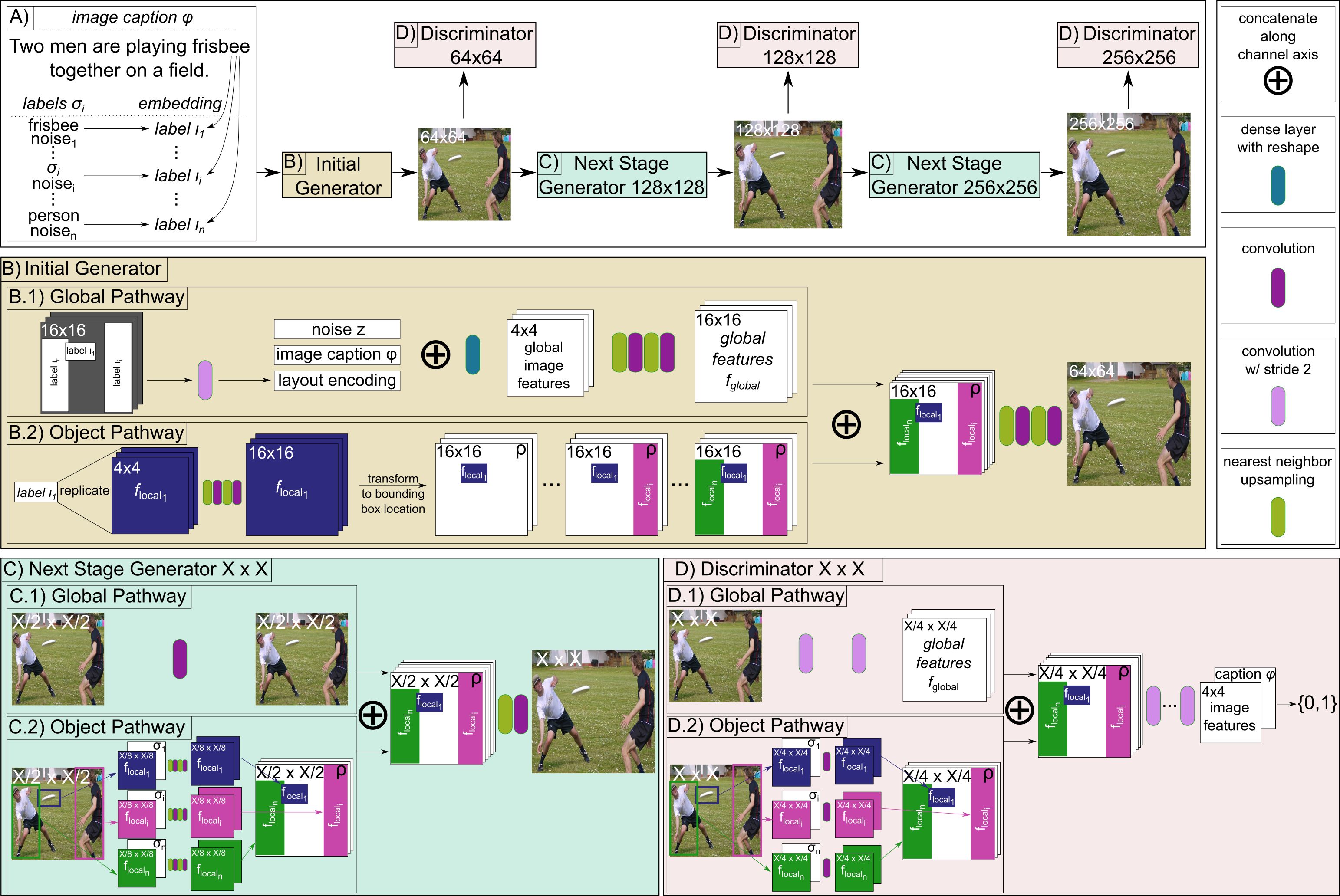}
	\caption{Overview of our model architecture called \textit{OP-GAN}. The top row shows a high-level summary of our architecture, while the bottom two rows show details of the individual generators and discriminators.}
	\label{fig:model}
	\end{figure*}
	
    A traditional generative adversarial network (GAN) \cite{goodfellow2014generative} consists of two networks: a generator $G$ which generates new data points from randomly sampled inputs, and a discriminator $D$ which tries to distinguish between generated and real data samples.
	In conditional GANs (cGANs) \cite{mirza2014conditional} both the discriminator and the generator are conditioned on additional information, e.g. a class label or textual information.
	This has been shown to improve performance and leads to more control over the data generating process.
	For a conventional cGAN with generator $G$, discriminator $D$, condition $c$ (e.g. a class label), data point $x$, and a randomly sampled noise vector $z$ the training objective $V$ is:
	\begin{equation}
	\begin{split}
	\underset{G}{\text{min}}\ \underset{D}{\text{max}}\ V(D,G) =
	\mathbb{E}_{(x,c)\sim p_{\text{data}}}[log\ D(x,c)] + \\
	\mathbb{E}_{(z)\sim p_{z}, (c)\sim p_{\text{data}}}[log(1-D(G(z,c),c))].
	\end{split}
	\label{eq:cGAN}
	\end{equation}
	
	We use the AttnGAN \cite{xu2017attngan} as our baseline architecture and add our object-centric modifications to it.
	The AttnGAN is a conditional GAN for text-to-image synthesis that uses attention and a novel additional loss to improve the quality of the generated images.
	It consists of a generator and three discriminators as shown in the top row of \autoref{fig:model}.
	Attention is used such that different words of the caption have more or less influence on different regions of the image.
	This means that, for example, the word ``sky'' has more influence on the generation of the top half of the image than the word ``grass'' even if both words are present in the image caption.
	
	\cite{xu2017attngan} also introduce the Deep Attentional Multimodal Similarity Model (DAMSM) which computes the similarity between images and captions.
	This DAMSM is used during training to provide additional, fine-grained feedback to the generator about how well the generated image matches its caption.
	We adapt the AttnGAN architecture with multiple object pathways which are learned end-to-end in both the discriminator and the generator, see \textit{B} and \textit{C} in \autoref{fig:model}.
	
	These object pathways are conditioned on individual object labels (e.g. ``person'', ``car'', etc.) and the same object pathway is applied multiple times at a given image resolution at different locations and for different objects.
	This is similar to the approach introduced by \cite{hinz2019generating}.
	However, \cite{hinz2019generating} only use one object pathway in the generator at a small resolution and only one discriminator was equipped with an object pathway.
	In our approach, the generator contains three object pathways at various resolutions ($16\times 16$, $64\times 64$, and $128\times 128$) to further refine object features at higher resolutions and each of our three discriminators is equipped with its own object pathway, see \textit{D} in \autoref{fig:model}.
	
	For a given image caption $\varphi$ we have several objects which are associated with this caption and which we represent with one-hot vectors $\sigma_i, i=1...n$ (e.g. $\sigma_0=\text{person}$, $\sigma_1=\text{car}$, etc.).
	Each object pathway at a given resolution is applied iteratively for each of the objects $\sigma_i$.
	The location is determined by a bounding box describing the object's location and size.
	Each object pathway starts with an ``empty'' zero-tensor $\rho$ and the features that are generated (generator) or extracted (discriminator) are added onto $\rho$ at the location of the specific object's bounding box.
	After the object pathway has processed each object, $\rho$ contains features at each object location and is zero everywhere else.
	
	For the generator, we first concatenate the image caption's embedding $\varphi$, the one-hot label $\sigma_i$, and a randomly sampled noise vector $z$.
	We use this concatenated vector to obtain the final conditioning label $\iota_i$ for the current object $\sigma_i$:
	\begin{equation}
	\label{eq:conditioning-label}
		\iota_i = \text{F}(\varphi, z, \sigma_i),
	\end{equation}
	where $F$ is a fully connected layer followed by a non-linearity (\textit{A} in \autoref{fig:model}).
	
	The generator's first object pathway (\textit{B.2} in \autoref{fig:model}) takes this conditioning label $\iota_i$ and uses it to generate features for the given object at a spatial resolution of $16\times 16$.
	The features are then transformed onto $\rho$ into the location of the respective bounding box with a spatial transformer network (STN) \cite{jaderberg2015spatial}.
	This procedure is repeated for each object $\sigma_i$ associated with the given caption $\varphi$.
	
	The global pathway in the first generator also gets the locations and labels $\iota_i$ for the individual objects.
	It spatially replicates these labels at the locations of the respective bounding boxes and then applies convolutional layers to the resulting layout to obtain a layout encoding (\textit{B.1} in \autoref{fig:model}).
	This layout encoding, the image caption $\varphi$, and the noise vector $z$ are used to generate coarse features for the image at a low resolution. 
	
	At higher levels in the generator, the object pathways are conditioned on the object features of the current object and the one-hot label $\sigma_i$ for that object (\textit{C.2} in \autoref{fig:model}).
	For this, we again use an STN to extract the features at the bounding box location of the object $\sigma_i$ and resize the features to a spatial resolution of $16\times 16$ (second object pathway) or $32\times 32$ (third object pathway).
	We obtain a conditioning label in the same manner as for the first object pathway (\autoref{eq:conditioning-label}), replicate it spatially to the same dimension as the extracted object features, and concatenate it with the object features along the channel axis.
	Following this, we apply multiple convolutional layers and upsampling to update the features of the given object.
	Finally, as in the first object pathway, we use an STN to transform the features into the bounding box location and add them onto $\rho$.
	The global pathway in the higher layers (\textit{C.1} in \autoref{fig:model}) stays unchanged from the baseline architecture \cite{xu2017attngan}.
	
	Our final loss function for the generator is the same as in the original AttnGAN and consists of an unconditional, a conditional, and a caption-image matching part.
	The unconditional loss is
	\begin{equation}
	\mathcal{L}_{G}^{\text{uncon}} =
	-\mathbb{E}_{(\hat{x})\sim p_{G}}[log\ D(\hat{x}))],
	\end{equation}
	the conditional loss is
	\begin{equation}
	\mathcal{L}_{G}^{\text{con}} =
	-\mathbb{E}_{(\hat{x})\sim p_{G}, (c)\sim p_{\text{data}}}[log\ D(\hat{x}, c))],
	\end{equation}
	and the caption-image matching loss is
	\begin{equation}
	\mathcal{L}_{G}^{\text{DAMSM}} =
	-\mathbb{E}_{(\hat{x})\sim p_{G}, (c)\sim p_{\text{data}}}[log\ D(\hat{x}, c))],
	\end{equation}
	which measures text-image similarity at the word level and is calculated with the pre-trained models provided by \cite{xu2017attngan}.
	The complete loss for the generator then is:
	\begin{equation}
	\mathcal{L}_{G} = \mathcal{L}_{G}^{\text{uncon}} + \mathcal{L}_{G}^{\text{con}} + \lambda \mathcal{L}_{G}^{\text{DAMSM}},
	\end{equation}
	where we set $\lambda = 50$ as in the original implementation.
	
	As in our baseline architecture, we employ three discriminators at three spatial resolutions: $64\times 64$, $128\times 128$, and $256\times 256$.
	Each discriminator possesses a global and an object pathway which extract features in parallel (\textit{D} in \autoref{fig:model}).
	In the object pathway we use an STN to extract the features of object $\sigma_i$ and  concatenate them with the one-hot vector $\sigma_i$ describing the object.
	The object pathway then applies multiple convolutional layers before adding the extracted features onto $\rho$ at the location of the bounding box.
	
	The global pathway in each of the discriminators works on the full input image and applies convolutional layers with stride two to decrease the spatial resolution (\textit{D.1}).
	Once the spatial resolution reaches that of the tensor $\rho$ we concatenate the two tensors (full image features and object features $\rho$) along the channel axis and use convolutional layers with stride two to further reduce the spatial dimension until we reach a resolution of $4\times 4$.
	
	We calculate both a conditional (image and image caption) and an unconditional (only image) loss for each of the discriminators.
	The conditional input $c$ during training consists of the image caption embedding $\varphi$ and the information about objects $\sigma_i$ (bounding boxes and object labels) associated with the image $x$, i.e. $c = \{\varphi, \sigma_i\}$.
	In the unconditional case the discriminators are trained to classify images as real or generated without any influence of the image caption by minimizing the following loss:
	\begin{equation}
	\mathcal{L}_{D_i}^{\text{uncon}} =
	-\mathbb{E}_{(x)\sim p_{\text{data}}}[log\ D(x)] - \\
	\mathbb{E}_{(\hat{x})\sim p_{G}}[log(1-D(\hat{x}))].
	\end{equation}
	In order to optimize the conditional loss we concatenate the extracted features with the image caption embedding $\varphi$ along the channel axis and minimize
	\begin{equation}
	\begin{split}
	\mathcal{L}_{D_i}^{\text{con}} =
	-\mathbb{E}_{(x, c)\sim p_{\text{data}}}[log\ D(x, c)] \\
	- \mathbb{E}_{(\hat{x})\sim p_{G}, (c)\sim p_{\text{data}}}[log(1-D(\hat{x}, c))].
	\end{split}
	\end{equation}
	for each discriminator.
	Finally, to specifically train the discriminators to check for caption-image consistency we use the matching aware discriminator loss \cite{reed2016generative} with mismatching caption-image pairs and minimize
	\begin{equation}
	\mathcal{L}_{D_i}^{\text{cls}} =
	-\mathbb{E}_{(x, \sigma)\sim p_{\text{data}}, (\varphi)\sim p_{\text{data}}}[log(1-D(x,c))],
	\end{equation}
	where image $x$ and caption $\varphi$ are sampled individually and randomly from the data distribution and are, therefore, unlikely to align.
	
	We introduce an additional loss term similar to the matching aware discriminator loss $V_{\text{cls}}(D)$ which works on individual objects.
	Instead of using mismatching image-caption pairs, we use correct image-caption pairs, but with incorrect bounding boxes and minimize:
	\begin{equation}
	\mathcal{L}_{D_i}^{\text{obj}} =
	-\mathbb{E}_{(x, \varphi)\sim p_{\text{data}}, (\sigma)\sim p_{\text{data}}}[log(1-D(x,c))].
	\end{equation}
	
	Thus, the complete objective we minimize for each individual discriminator is:
	\begin{equation}
	\mathcal{L}_{D_i} = \mathcal{L}_{D_i}^{\text{uncon}} + \mathcal{L}_{D_i}^{\text{con}} + \mathcal{L}_{D_i}^{\text{cls}} + \mathcal{L}_{D_i}^{\text{obj}}.
	\end{equation}
	We leave all other training parameters as in the original implementation \cite{xu2017attngan} and the training procedure itself also stays the same.

\section{Evaluation of Text-to-Image Models}
\label{sec:eval-metric}	
    Quantitatively evaluating generative models is difficult \cite{theis2016note}.
	While there are several evaluation metrics that are commonly used to evaluate GANs, many of them have known weaknesses and are not designed specifically for text-to-image synthesis tasks.
	In the following, we first discuss some of the common evaluation metrics for GANs, their weaknesses, and why they might be inadequate for evaluating text-to-image synthesis models.
	Following this, we introduce our novel evaluation metric, Semantic Object Accuracy (SOA), and describe how it can be used to evaluate text-to-image models in more detail.
	
	\subsection*{Current Evaluation Metrics}	
	\textbf{Inception Score and Fr\'{e}chet Inception Distance}\ \ Most GAN approaches are trained on relatively simple images which only contain one object at the center (e.g. ImageNet, CelebA, etc).
	These methods are evaluated with metrics such as the Inception Score (IS) \cite{salimans2016improved} and Fr\'{e}chet Inception Distance (FID) \cite{heusel2017gans}, which use an Inception-Net usually pre-trained on ImageNet.
	The IS evaluates roughly how distinctive an object in each image is (i.e. ideally the classification layer of the Inception-Net has small entropy) and how many different objects the GAN generates overall (i.e. high entropy in the output of different images).
	The FID measures how similar generated images are to a control set of images, usually the validation set by calculating the distance in feature space between generated and real images.
	Consequently, the IS should be as high as possible, while the FID should be as small as possible.
	
\begin{figure}[!t]
		\centering
		\includegraphics[width=0.48\textwidth]{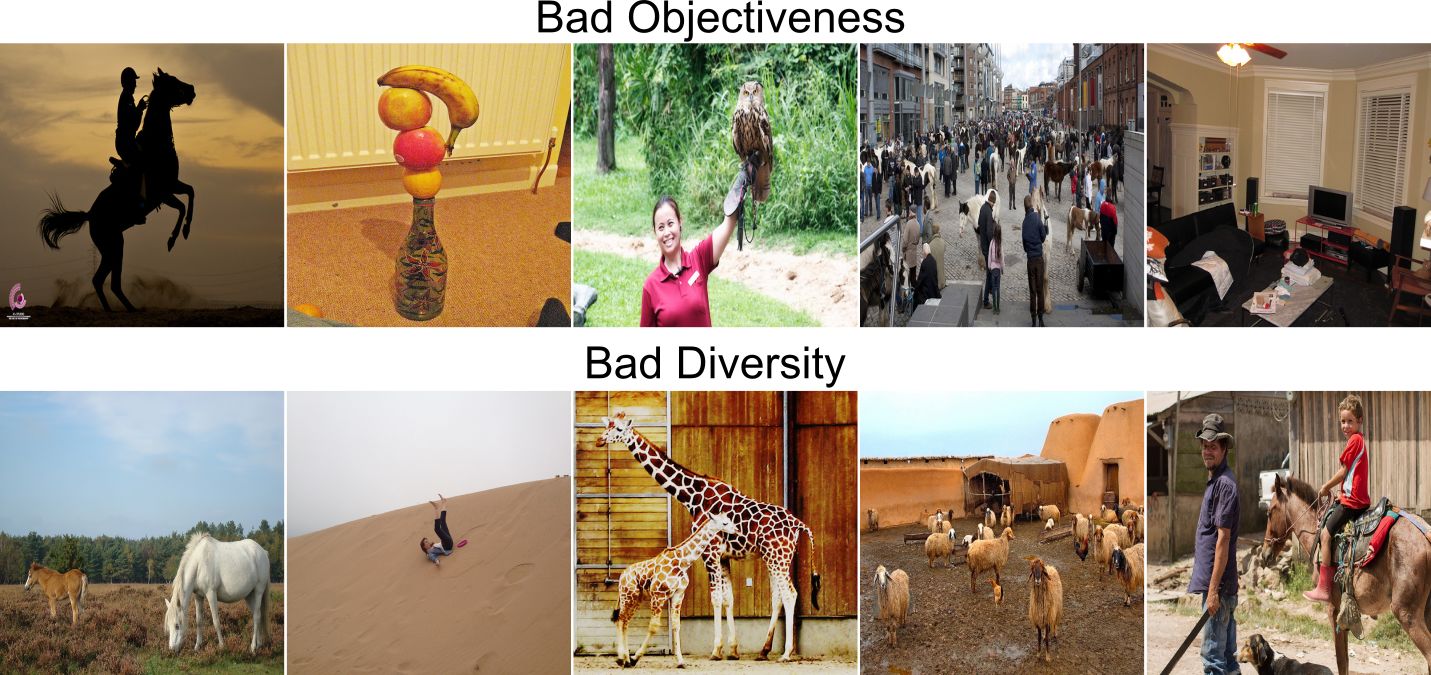}
		\caption{Examples when IS fails for COCO images. The top row shows images for which the Inception-Net has very high entropy in its output layer, possibly because the images contain more than one object and are often not centered. The second row shows images containing different objects and scenes which were nonetheless all assigned to the same class by the Inception-Net, thereby negatively affecting the overall predicted diversity in the images.}
		\label{fig:metric:IS}
	\end{figure}

	Both evaluation metrics have known weaknesses \cite{borji2019pros, barratt2018note}.
	For example, the IS does not measure the similarity between objects of the same class, so a network that only generates one ``perfect'' sample for each class can achieve a very good IS despite showing an intra-class mode dropping behavior.
	Li et al. \cite{li2019object} also note that the IS overfits within the context of text-to-image synthesis and can be ``gamed'' by increasing the batch size at the end of the training.
	Furthermore, the IS uses the output of the classification layer of an Inception-Net pre-trained on the ImageNet data set.
	This might not be the best approach for a more complex data set in which each image contains multiple objects at distinct locations throughout the image, as opposed to the ImageNet data set which consists of images usually depicting one object in the image center.
	\autoref{fig:metric:IS} shows some exemplary failure cases of the IS on images sampled from the COCO data set.
	
	The FID relies on representative ground truth data to compare the generated data against and also assumes that features are of Gaussian distribution, which is often not the case.
	For more complex data sets the FID also still suffers from the problem that the image statistics are obtained with a network pre-trained on ImageNet which might not be a representative data set.	
	Finally, neither the IS nor the FID take the image caption into account during their evaluation.
	
	\textbf{VS similarity and R-precision}\ \ \cite{zhang2018photographic} introduce the visual-semantic similarity (VS similarity) metric which measures the distance between a generated image and its caption.
	Two models are trained to embed images and captions respectively and then minimize the cosine distance between embeddings of matching image-caption pairs while maximizing the cosine distance between mismatching image-caption pairs.
	A good model then achieves high VS similarity between a generated image and its associated caption.	
	
	\cite{xu2017attngan} use the R-precision metric to evaluate how well an image matches a given description or caption.
	The R-precision score is similar to VS similarity, but instead of scoring the VS similarity between a given image and caption it instead performs a ranking of the similarity between the real caption and randomly sampled captions for a given generated image.
	For this, first, an image is generated conditioned on a given caption.
	Then, another 99 captions are chosen randomly from the data set.
	Both the generated images and the 100 captions are then encoded with the respective image and text encoder.
	Similar to VS similarity the cosine distance between the image embedding and each caption embedding is used as proxy for the similarity between the given image and caption.
	The 100 captions are then ordered in descending similarity and the top \textit{k} (usually \textit{k}=1) most similar captions are used to calculate the R-precision.
	Intuitively, R-precision calculates if the real caption is more similar to the generated image (in feature space) than 99 randomly sampled captions.
	
\begin{figure}[!t]
		\centering
		\includegraphics[width=0.48\textwidth]{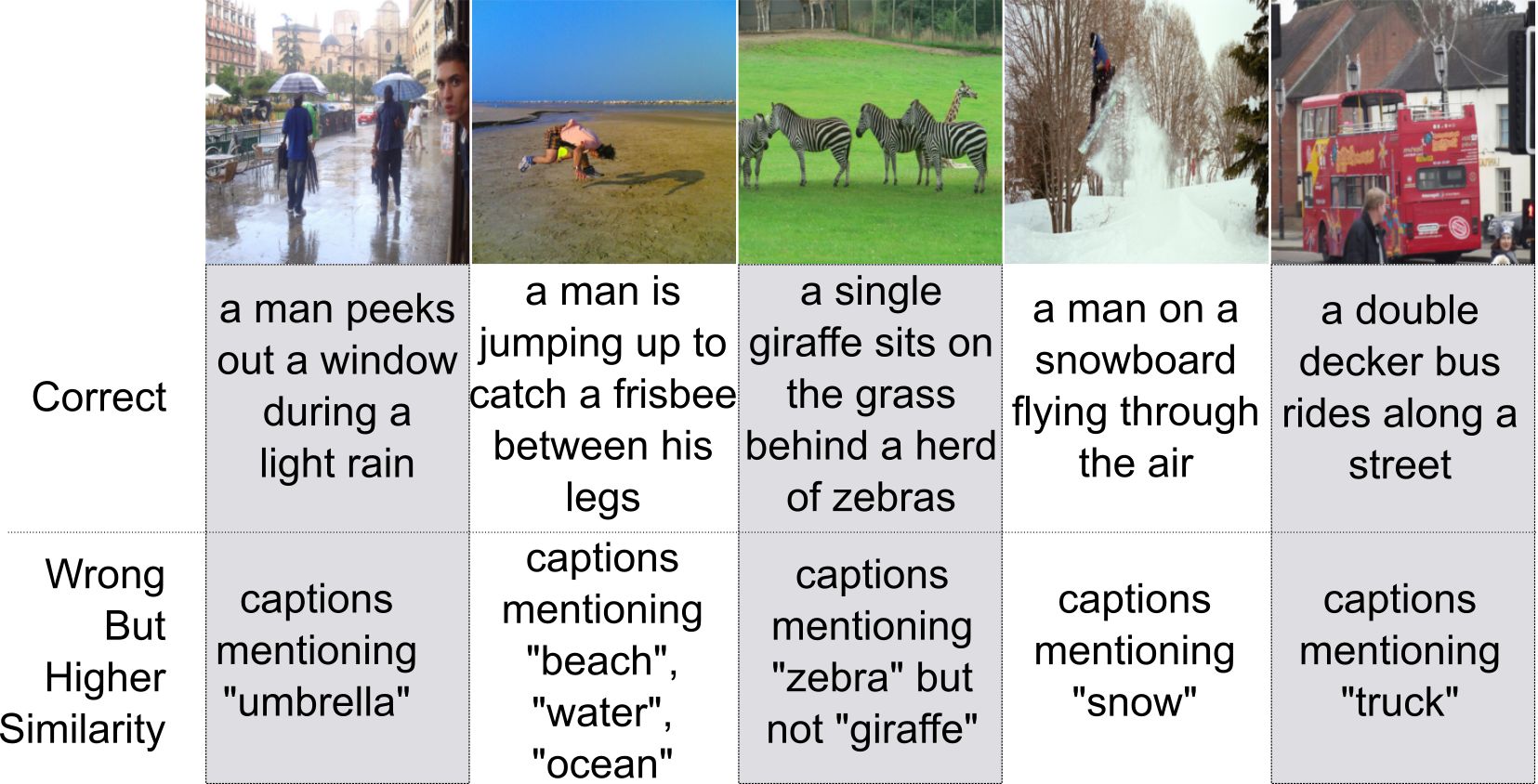}
		\caption{Examples when R-precision fails for COCO images. The top row shows images from the COCO data set. The middle row shows the correct caption and the bottom row gives examples for characteristics of captions that are rated as being more similar than the original caption.}
		\label{fig:metric:R-prec}
	\end{figure}

	The drawback of both metrics is that they do not evaluate the quality of individual objects.
	For example the real caption could state that \textit{``a person stands on a snowy hill''} while the 99 random captions do not mention ``snow'' (which usually covers most of the background in the generated image) or ``person'' (but e.g. giraffe, car, bedroom, etc).
	In this case, an image with only white background (snow) would already make the real caption rank very highly in the R-precision metric and achieve a high VS similarity.
	See \autoref{fig:metric:R-prec} for a visualization of this.
	As such, this metric does not focus on the quality of individual objects but rather concentrates on global background and salient features.
	
	\textbf{Classification Accuracy Score}\ \ \cite{ravuri2019classification} introduce the Classification Accuracy Score (CAS) to evaluate conditional image generation models, similar to \cite{shmelkov2018good}.
	For this, a classifier is trained on images generated by the conditional generative model.
	The classifier's performance is then evaluated on the original test set of the data set that was used to train the generative model.
	If the classifier achieves high accuracy on the test set, this indicates that the data it was trained on is representative of the real distribution.
	The authors find that neither the IS, the FID, nor combinations thereof are predictive of the CAS, further indicating that the IS and FID are only of limited use for evaluating image quality.
	
	\textbf{Caption Generation}\ \ \cite{hong2018inferring} suggest evaluating text-to-image models by comparing original captions with captions obtained from generated images.
	The intuition is that if the generated image is relevant to its caption, then it should be possible to infer the original text from it.
	To this end, \cite{hong2018inferring} use a pre-trained caption generator \cite{vinyals2016show} to generate captions for each synthesized image and compare these to the original ones through standard language similarity metrics, i.e. BLEU, METEOR, and CIDEr.
	Except for CIDEr, these metrics were originally developed to evaluate machine translation and text summarization methods and were only later adopted for the evaluation of image captions.
	
	One challenge with this caption generation approach is that often many different captions are valid for a given image.
	Even if two captions are not similar, this does not necessarily imply that they do not describe the same image \cite{anderson2016spice}.
	Furthermore, it has been shown that metrics such as BLEU, METEOR, and CIDEr are primarily sensitive to n-gram overlap which is neither necessary nor sufficient for two sentences to convey the same meaning \cite{gimenez2007linguistic,anderson2016spice,madhyastha2019vifidel} and do also not necessarily correlate with human judgments of captions \cite{vinyals2016show,kilickaya2017re}.
	Finally, there is no requirement that captions, either real or generated, need to focus on specific objects.
	Instead, captions can also describe the general layout of a given scene (e.g. \textit{a busy street with lots of traffic}) without explicitly mentioning specific objects.
	Some of these limitations might potentially be overcome in the future by novel image caption evaluation metrics that focus more on objects and semantic content in the scene \cite{anderson2016spice,madhyastha2019vifidel,agarwal2020egoshots}.
	
	\textbf{Other Approaches}\ \ In contrast to the IS, which measures the diversity of a whole set of images, the diversity score \cite{zhao2018image} measures the perceptual difference between a pair of images in feature space.
	This metric can be useful when images are generated from conditional inputs (e.g. labels or scene layouts) to examine whether a model can generate diverse outputs for a given condition.
	However, the metric does not convey anything directly about the quality of the generated images or their congruence with any conditional information.
	\cite{chen2017photographic, wang2018high, park2019semantic} run a semantic segmentation network on generated images and compare the predicted segmentation mask to the ground truth segmentation mask used as input for the model.
	However, this metric needs a ground truth semantic segmentation mask and does not provide information about specific objects within the image.
	
	\subsection*{Semantic Object Accuracy (SOA)}	
	So far, most evaluation metrics are designed to evaluate the holistic image quality but do not evaluate individual areas or objects within an image.
	Furthermore, except for \textit{Caption Generation} and \textit{R-precision}, none of the scores take the image caption into account when evaluating generated images.
	To address the challenges and issues mentioned above we introduce a novel evaluation metric based on a pre-trained object detection network.\footnote{Code for the evaluation metric and all experiments: \url{https://github.com/tohinz/semantic-object-accuracy-for-generative-text-to-image-synthesis}}
	The pre-trained object detector evaluates images by checking if it recognizes objects that the image should contain based on the caption.
	For example, if the image caption is \textit{``a person is eating a pizza''} we can infer that the image should contain both a person and a pizza and the object detector should be able to recognize both objects within the image.
	Since this evaluation measures directly whether objects specifically mentioned in the caption are recognizable in an image we call this metric \textit{Semantic Object Accuracy} (SOA).
	
	Some previous works have used similar approaches to evaluate the quality of the generated images.
	\cite{hinz2019generating} evaluate how often expected objects (based on the caption) are detected by an object detector.
	However, only a subset of the captions is evaluated and the evaluated captions contain false positives (e.g. captions containing the phrase ``hot dog'' are evaluated based on the assumption that the image should contain a dog).
	\cite{sah2018semantically} introduce a detection score that calculates (roughly) whether a pre-trained object detector detects an object in a generated image with high certainty.
	However, no information from the caption is taken into account, meaning any detection with high confidence is ``good'' even if the detected object does not make sense in the context of the caption.
	\cite{deng2018probabilistic} use a pre-trained object detector to calculate the mean average precision and report precision-recall curves.
	However, the evaluation is done on synthetic data sets and without textual information as conditional input.
	\cite{zhao2018image} use classification accuracy as an evaluation metric in which they report the object classification accuracy in generated images.
	For this, they use a ResNet-101 model which is trained on real objects cropped and resized from the original data.
	However, in order to calculate the score, the size and location of each object in the generated image must be known, so this evaluation is not directly applicable to approaches that do not use scene layouts or similar representations.
	\cite{vo2019visual} use recall and intersection-over-union (IoU) to evaluate the bounding boxes in their generated scene layout but do not apply these evaluations to generated images directly.
	
	\textbf{SOA}\ \ Since we work with the COCO data set we filter all captions in the validation set for specific keywords that are related to the available labels for objects (e.g. person, car, zebra, etc).
	For each of the 80 available labels in the COCO data set we find all captions that imply the existence of the respective object and generate three images for each of the captions.
	The supplementary material gives a detailed overview of how exactly the captions were chosen for each label.
	We then run the YOLOv3 network \cite{redmon2018yolov3} pre-trained on the COCO data set on each of the generated images and check whether it recognizes the given object.
	We report the recall as a class average (SOA-C), i.e. in how many images per class the YOLOv3 on average detects the given object, and as an image average (SOA-I), i.e. on average in how many images a desired object was detected.
	Specifically, the SOA-C is calculated as
	\begin{equation}
	\text{SOA-C} = \frac{1}{\vert C\vert}\sum_{c\in C}\frac{1}{\vert I_c\vert}\sum_{i_c\in I_c} \text{YOLOv3}(i_c),
	\end{equation}
	for object classes $c\in C$ and images $i\in I_c$ that are supposed to contain an object of class $c$.
	The SOA-I is calculated as
	\begin{equation}
	\text{SOA-I} = \frac{1}{\sum_{c\in C}\vert I_c\vert}\sum_{c\in C}\sum_{i_c\in I_c} \text{YOLOv3}(i_c),
	\end{equation}
	and
	\begin{equation}
	\text{YOLOv3}(i_c) = 
	\begin{cases}
	1 & \!\text{\hspace{-0.15cm}if YOLOv3 detected an object of class }c\\
	0 & \!\text{\hspace{-0.15cm}otherwise}
	\end{cases}.
	\end{equation}
	Since many images can also contain objects that are not specifically mentioned (for example an image described by \textit{``lots of cars are on the street''} could still contain persons, dogs, etc) in the caption we do not calculate a false negative rate but instead only focus on the recall, i.e. the true positives.
	
	\textbf{SOA-Intersection over Union}\ \ Several approaches (e.g. \cite{hinz2019generating, li2019object, hong2018inferring, zhao2018image, vo2019visual}) use additional conditioning information such as scene layouts or bounding boxes.
	For these approaches, our evaluation metric can also calculate the intersection over union (IoU) between the location at which different objects should be and locations at which they are detected, which we call SOA-IoU.
	To calculate the IoU we use every image in which the YOLOv3 network detected the respective object.
	Since many images contain multiple instances of a given object we calculate the IoU between each predicted bounding box for the given object and each ground truth bounding box.
	The final IoU for a given image and object is then the maximum of the values, i.e. the reported IoU is an upper bound on the actual IoU.
	
	Overall this approach allows a more fine-grained evaluation of the image content since we can now focus on individual objects and their features.
	To get a better idea of the overall performance of a model we calculate both the class average recall/IoU (SOA-C/SOA-IoU-C) and image average recall/IoU (SOA-I/SOA-IoU-I).
	Additionally, we report the SOA-C for the forty most and least common labels (SOA-C-Top40 and SOA-C-Bot40) to see how well the model can generate objects of common and less common classes.
	
\section{Experiments}	
	\begin{table*}\centering
		\caption{Inception Score (IS), Fr\'{e}chet Inception Distance (FID), R-precision, Caption Generation with CIDEr, and Semantic Object Accuracy on Class (SOA-C) and Image Average (SOA-I) on the MS-COCO data set. Results of our models are obtained with generated bounding boxes. Scores for models marked with $^\dag$ were calculated with a pre-trained model provided by the respective authors.}
		\label{tab:coco:results}
		\begin{tabular}{l c c c c c c }
			\toprule
			Model & IS $\uparrow$ & FID $\downarrow$ & R-precision (k=1) $\uparrow$ & CIDEr $\uparrow$ & SOA-C $\uparrow$ & SOA-I $\uparrow$ \\
			\midrule
			Original Images & $34.88 \pm 0.01$ & $6.09 \pm 0.05$ & $68.58 \pm 0.08$ & $0.795 \pm 0.003 $ & $74.97$ & $80.84$ \\
			\midrule
			AttnGAN \cite{xu2017attngan}$^\dag$ & $23.61 \pm 0.21$ & $33.10 \pm 0.11$ & $83.80$ & $0.695 \pm 0.005$ & $25.88$ & $39.01$ \\
			\rowcolor{Gray}\cite{huang2019realistic} & $23.74 \pm 0.36$ & & $86.44 \pm 3.38$ & & & \\
			ControlGAN \cite{li2019controllable} & $24.06 \pm 0.60$ & & $82.43$ \\
			\rowcolor{Gray}AttnGAN + OP \cite{hinz2019generating}$^\dag$ & $24.76 \pm 0.43$ & $33.35 \pm 1.15$ & $82.44$ & $0.689 \pm 0.008$ & $25.46$ & $40.48$ \\
			MirrorGAN \cite{qiao2019mirrorgan} & $26.47 \pm 0.41$ & & $74.52$  \\
			\rowcolor{Gray}Obj-GAN \cite{li2019object}$^\dag$ & $24.09 \pm 0.28$ & $36.52 \pm 0.13$ & $87.84 \pm 0.08$ & $0.783 \pm 0.002$ & $27.14$ & $41.24$ \\
			HfGAN \cite{huang2019hierarchically} & $27.53 \pm 0.25$ & & & \\
			\rowcolor{Gray}DM-GAN \cite{zhu2019dm}$^\dag$ & $32.32 \pm 0.23$ & $27.34 \pm 0.11$ & $\boldsymbol{91.87 \pm 0.28}$ & $\boldsymbol{0.823 \pm 0.002}$ & $33.44$ & $ 48.03 $ \\
			SD-GAN \cite{yin2019semantics} & $\boldsymbol{35.69 \pm 0.50}$ &  &  & \\
			\rowcolor{Gray}\textit{OP-GAN} (Best Model) & $27.88 \pm 0.12$ & $\boldsymbol{24.70 \pm 0.09}$ & $89.01 \pm 0.26$ & $0.819 \pm 0.004$ & $\boldsymbol{35.85}$ & $\boldsymbol{50.47}$ \\
			\midrule
			\textit{OPv2}, 0 obj & $26.80 \pm 1.01$ & $30.01 \pm 1.81$ & $83.87 \pm 1.22$ & $0.760 \pm 0.004$ & $26.04 \pm 1.47$ & $37.56 \pm 1.27$ \\
			\textit{OPv2}, 1 obj & $27.68 \pm 0.47$ & $26.18 \pm 0.27$ & $87.37 \pm 0.60$ & $0.798 \pm 0.013$ \\
			\textit{OPv2}, 3 obj & $27.78 \pm 0.50$ & $26.45 \pm 0.40$ & $87.74 \pm 1.08$ & $0.805 \pm 0.011$ \\
			\textit{OPv2}, 10 obj & $27.66 \pm 0.34$ & $26.52 \pm 0.44$ & $87.73 \pm 0.98$  & $0.806 \pm 0.006$ & $33.82 \pm 0.69$ & $48.39 \pm 1.01$ \\
			\rowcolor{Gray}\textit{OPv2} + \textit{BBL}, 0 obj & $24.60 \pm 1.25$ & $33.03 \pm 0.76$ & $81.27 \pm 1.45$ & $0.735 \pm 0.029$ & $24.00 \pm 2.13$ & $34.01 \pm 2.89$ \\
			\rowcolor{Gray}\textit{OPv2} + \textit{BBL}, 1 obj & $26.34 \pm 0.55$ & $26.59 \pm 1.04$ & $86.42 \pm 0.60$ & $0.783 \pm 0.006$ & & \\
			\rowcolor{Gray}\textit{OPv2} + \textit{BBL}, 3 obj & $26.52 \pm 0.47$ & $26.74 \pm 1.08$ & $87.08 \pm 0.60$ & $0.793 \pm 0.013$ & & \\
			\rowcolor{Gray}\textit{OPv2} + \textit{BBL}, 10 obj & $26.48 \pm 0.58$ & $26.83 \pm 1.10$ & $86.80 \pm 0.56$ & $0.794 \pm 0.015$ & $33.19 \pm 0.40$ & $48.24 \pm 0.68$ \\
			\textit{OPv2} + \textit{MO}, 0 obj & $24.32 \pm 1.65$ & $35.36 \pm 1.95$ & $79.75 \pm 1.87$ & $0.695 \pm 0.015$ & $21.15 \pm 1.47$ & $30.24 \pm 2.36$ \\
			\textit{OPv2} + \textit{MO}, 1 obj & $27.36 \pm 0.49$ & $25.06 \pm 1.11$ & $88.33 \pm 0.81$ & $0.789 \pm 0.008$ \\
			\textit{OPv2} + \textit{MO}, 3 obj & $27.65 \pm 0.37$ & $24.96 \pm 1.12$ & $ 89.13 \pm 0.42 $ & $0.807 \pm 0.014$ \\
			\textit{OPv2} + \textit{MO}, 10 obj & $27.59 \pm 0.43$ & $\boldsymbol{24.94 \pm 1.09}$ & $\boldsymbol{89.14 \pm 0.41}$ & $0.805 \pm 0.013$ & $33.46 \pm 1.01$ & $47.93 \pm 1.56$ \\
			\rowcolor{Gray}\textit{OPv2} + \textit{BBL} + \textit{MO}, 0 obj & $21.84 \pm 0.83$ & $45.79 \pm 1.16$ & $72.71 \pm 1.75$ & $0.626 \pm 0.025$ & $16.55 \pm 1.81$ & $22.76 \pm 2.17$ \\
			\rowcolor{Gray}\textit{OPv2} + \textit{BBL} + \textit{MO}, 1 obj & $27.61 \pm 0.67$ & $26.19 \pm 0.82$ & $87.85 \pm 0.25$ & $0.791 \pm 0.009$ & & \\
			\rowcolor{Gray}\textit{OPv2} + \textit{BBL} + \textit{MO}, 3 obj & $\boldsymbol{28.04 \pm 0.65}$ & $25.91 \pm 1.03$ & $88.90 \pm 0.24$ & $0.810 \pm 0.009$ & & \\
			\rowcolor{Gray}\textit{OPv2} + \textit{BBL} + \textit{MO}, 10 obj & $27.90 \pm 0.79$ & $ 25.80 \pm 1.01 $ & $89.00 \pm 0.17$ & $\boldsymbol{0.814 \pm 0.007}$ & $\boldsymbol{34.51 \pm 1.12}$ & $\boldsymbol{48.90 \pm 0.72}$ \\
			\bottomrule
		\end{tabular}
	\end{table*}
	
	We perform multiple experiments and ablation studies.
	In a first step, we add the object pathway (OP) on multiple layers of the generator and to each discriminator and call this model \textit{OPv2}.
	We also train this model with the additional bounding box loss we introduced in \autoref{sec:approach}.
	When the model is trained with the additional bounding box loss we refer to it as \textit{BBL}. 
	
	Different approaches differ in how many objects per image are used during training.
	If an image layout is used, typically all objects (foreground and background) are used as conditioning information.
	Other approaches limit the number of objects during per training \cite{johnson2018image, hinz2019generating}.
	To examine the effect of training with different numbers of objects per image we train our approach with either a maximum of three objects per image (standard) or with up to ten objects per image, which we refer to as many objects (\textit{MO}).
	When training with a maximum of three objects per image we sample randomly from the training set at train time, i.e. each batch contains images which contain zero to three objects.
	If an image contains more than three objects we choose the three largest ones in terms of area of the bounding box.
	When training with up to ten objects per image we slightly change our sampling strategy so that each batch consists of images that contain the same amount of objects.
	This means that, e.g., each image in a batch contains exactly four objects, while in the next batch each image might contain exactly seven objects.
	This increases the training efficiency as most of the images contain fewer than five objects.
	
	As a result of the different settings we perform the following experiments:	
	\begin{enumerate}
		\item \textit{OPv2}: apply the object pathway (OP) on multiple layers of the generator and on all discriminators, training without the bounding box loss and with a maximum of three objects per image.
		\item \textit{OPv2} + \textit{BBL}: same as \textit{OPv2} but with the bounding box loss added to the discriminator loss term.
		\item \textit{OPv2} + \textit{MO}: same as \textit{OPv2} but with a maximum of ten objects per image.
		\item \textit{OPv2} + \textit{BBL} + \textit{MO} (\textit{OP-GAN}): combination of all three approaches.
	\end{enumerate}
	We train each model three times on the 2014 split of the COCO data set.
	At test time we use bounding boxes generated by a network \cite{li2019object} as the conditioning information.
	Therefore, except for the image caption no other ground truth information is used at test time.

\section{Evaluation and Analysis}	
	\autoref{tab:coco:results} and \autoref{tab:coco:soa} give an overview of our results for the COCO data set.
	The first half of the table shows the results on the original images from the data set and from related literature while the second half shows our results.
	To make a direct comparison we calculated the IS, FID, CIDEr, and R-precision scores ourselves for all models which are provided by the authors.
	As such, the values from AttnGAN \cite{xu2017attngan}, AttnGAN+OP \cite{hinz2019generating}, Obj-GAN \cite{li2019object}, and DM-GAN \cite{zhu2019dm} are the ones most directly comparable to our reported values since they were calculated in the same way.
	
	Note that there is some inconsistency in how the FID is calculated in prior works.
	Some approaches, e.g. \cite{li2019object}, compare the statistics of the generated images only with the statistics of the respective ``original'' images (i.e. images corresponding to the captions that were used to generate a given image).
	We, on the other hand, generate 30,000 images from 30,000 randomly sampled captions and compare their statistics with the statistics of the full validation set.
	Many of the recent publications also do not report the FID or R-precision.
	This makes a direct comparison difficult as we show that the IS is likely the least meaningful score of the three since it easily overfits \cite{li2019object} and due to the reasons mentioned in \autoref{sec:eval-metric}.
	We calculate each of the reported values of our models  three times for each trained model (nine times in total) and report the average and standard deviation.
	To calculate the SOA scores we generate three images for each caption in the given class, except for the ``person'' class, for which we randomly sample 30,000 captions (from over 60,000) and generate one image for each of the 30,000 captions.
	
	\subsection*{Quantitative Results}	
	\begin{table*}[t]\centering
		\caption{Comparison of the recall values for the different models. We used generated bounding boxes to calculate the values.\newline Numbers in brackets show scores when the object pathway was not used at test time.}
		\label{tab:coco:soa}
		\begin{tabular}{l c c c c}
			\toprule
			Model & SOA-C / IoU & SOA-I / IoU & SOA-C-Top40 / IoU & SOA-C-Bot40 / IoU \\
			\midrule
			Original Images & $74.97$ / $0.550$ & $80.84$ / $0.570$ & $78.77$ / $0.546$ & $71.18$ / $0.554$ \\
			AttnGAN \cite{xu2017attngan} & $25.88$ / $\ \ --\ $ & $39.01$ / $\ \ --\ $ & $37.47$ / $\ \ --\ $ & $14.29$ / $\ \ --\ $ \\
			AttnGAN + OP \cite{hinz2019generating} & $25.46$ / $0.236$ & $40.48$ / $0.311$ & $39.77$ / $0.308$ & $11.15$ / $0.164$ \\
			Obj-GAN \cite{li2019object} & $27.14$ / $0.513$ & $41.24$ / $0.598$ & $39.88$ / $0.587$ & $14.40$ / $0.438$ \\
			DM-GAN \cite{zhu2019dm} & $33.44$ / $\ \ --\ $ & $48.03$ / $\ \ --\ $ & $47.73$ / $\ \ --\ $ & $19.15$ / $\ \ --\ $ \\
			\midrule
			\textit{OPv2} & $33.82$ ($26.04$) / $0.207$ & $48.39$ ($37.56$) / $0.270$ & $48.34$ ($36.53$) / $0.260$ & $19.31$ ($15.55$) / $0.152$ \\
			\textit{OPv2} + \textit{BBL} & $33.19$ ($24.00$) / $0.210$ & $48.24$ ($34.01$) / $0.270$ & $47.96$ ($32.96$) / $0.261$ & $18.43$ ($15.04$) / $0.159$ \\
			\textit{OPv2} + \textit{MO} & $33.46$ ($21.15$) / $0.214$ & $47.93$ ($30.24$) / $0.275$ &  $47.84$ ($28.15$) / $0.264$ & $19.07$ ($14.15$) / $0.163$ \\
			\textit{OPv2} + \textit{BBL} + \textit{MO} & $34.51$ ($16.55$) / $0.217$ & $48.90$ ($22.76$) / $0.278$ & $49.70$ ($22.19$) / $0.269$ & $19.32$ ($10.91$) / $0.165$ \\
			\bottomrule
		\end{tabular}
	\end{table*}
	
    \textbf{Overall Results}\ \ As \autoref{tab:coco:results} shows, all our models outperform the baseline AttnGAN in all metrics.
	The IS is improved by $16-19\%$, the R-precision by $6-7\%$, the SOA-C by $28-33\%$, the SOA-I by $22-25\%$, the FID by $20-25\%$, and CIDEr by $15-18\%$.
	This was achieved by adding our object pathways to the baseline model without any further tuning of the architecture, hyperparameters, or the training procedure.
	Our approach also outperforms all other approaches based on FID, SOA-C, and SOA-I.
	While there are two approaches that report a IS higher than our models, it has previously been observed that this score is likely the least meaningful for this task and can be gamed to achieve higher numbers \cite{barratt2018note, li2019object}.
	Our user study also shows that the IS is the score that has the least predictive value for human evaluation.
	
	We also calculated each score using the original images of the COCO data set.
	For the IS we sampled three times 30,000 images from the validation set and resized them to $256\times 256$ pixels.
	These images were also used to calculate the CIDEr score.
	To calculate the FID we randomly sampled three times 30,000 images from the training set and compared them to the statistics of the validation set.
	The R-precision was calculated on three times 30,000 randomly sampled images and the corresponding caption from the validation set and the SOA-C and SOA-I were calculated on the real images corresponding to the originally chosen captions.
	
	As we can see, the IS is close to the current state of the art models with a value of $34.88$.
	It is possible to achieve a much higher IS on other, simpler data sets, e.g. IS $>100$ on the ImageNet data set \cite{brock2018large}.
	This indicates that the IS is indeed not a good evaluation metric, especially for complex images consisting of multiple objects and various locations.
	The difference between the R-precision on real and generated images is even larger.
	On the original images, the R-precision score is only $68.58$, which is much worse than what current models can achieve ($>88$).
	
	One reason for this might be that the R-precision calculates the cosine similarity between an image embedding and a caption embedding and measures how often the caption that was used to generate an image is more similar than 99 other, randomly sampled captions.
	However, the same encoders that are used to calculate the R-precision are also used during training to minimize the cosine similarity between an image and the caption it was generated from.
	As a result, the model might already overfit to this metric through the training procedure.
	Our observation is that the models tend to heavily focus on the background to make it match a specific word in the caption (e.g. images tend to be very white when the caption mentions ``snow'' or ``ski'', very blue when the caption mentions ``surf'' or ``beach'', very green when the caption mentions ``grass'' or ``savanna'', etc.)
	This matching might lead to a high R-precision score since it leads, on average, to a large cosine similarity.
	Real images do not always reflect this, since a large part of the image might be occupied by a person or an animal, essentially ``blocking out'' the background information.
	We see a similar trend for the CIDEr evaluation where many models achieve a score similar to the score reached by real images.
	Regardless of what the actual reason is, the question remains whether evaluation metrics like the IS, R-precision, and CIDEr are meaning- and helpful when models that can not (as of now) generate images that would be confused as ``real'' achieve scores comparable to or better than real images.
	
	\begin{figure*}[!t]
	\centering
	\includegraphics[width=\textwidth]{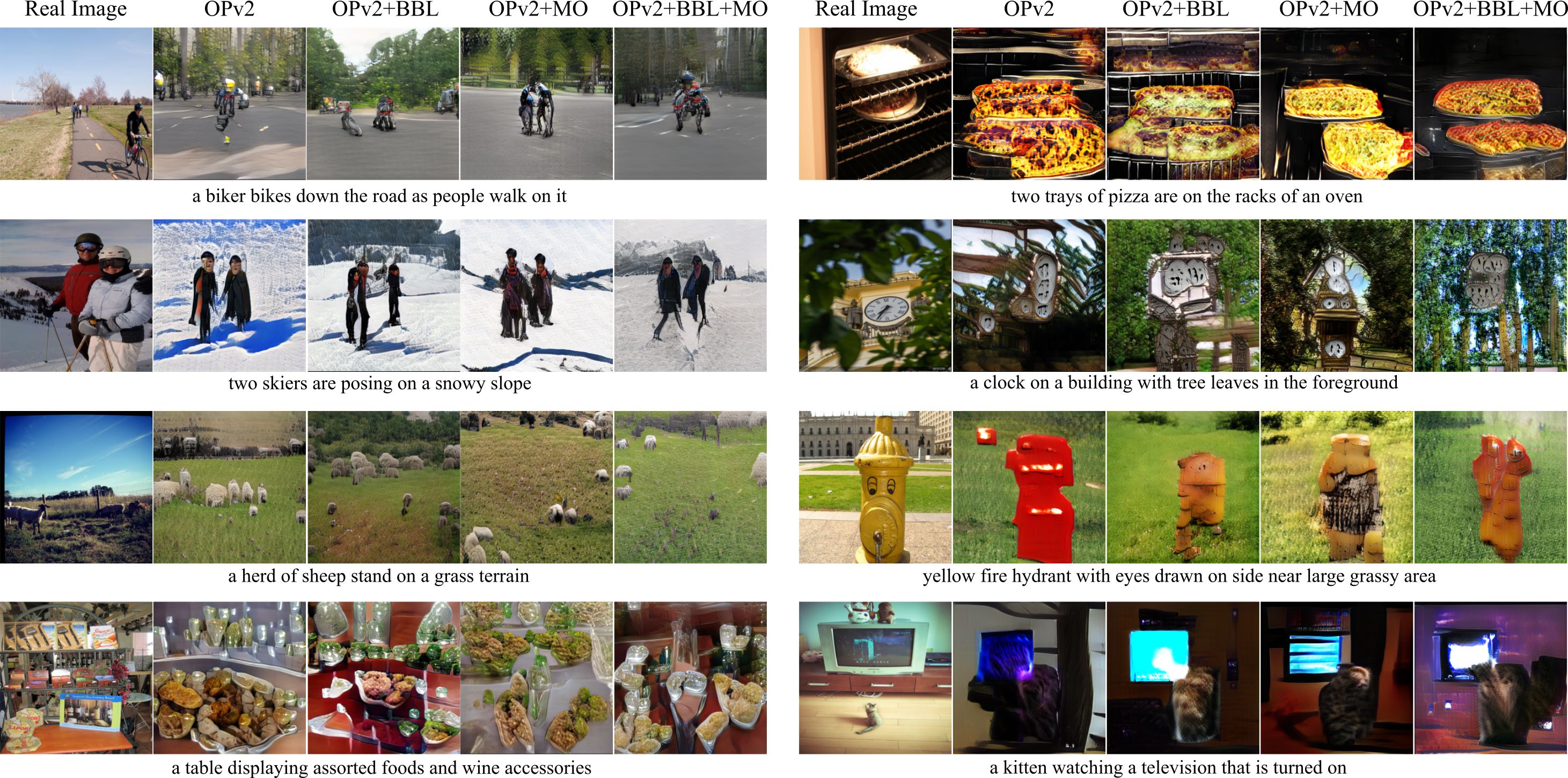}
	\caption{Comparison of images generated by different variations of our models.}
	\label{fig:comparison-our-models}
	\end{figure*}
	
	The FID and the SOA values are the only two evaluation metrics (that we used) for which none of the current state of the art models can come close to the values obtained with the original images.
	The FID is still much smaller on the real data ($6.09$) compared to what current models can achieve ($>24$ for the best models).
	While the FID still uses a network pre-trained on ImageNet it compares activations of convolutional layers for different images and is, therefore, likely still more meaningful and less dependent on specific object settings than the IS.
	Similarly, the SOA-C (SOA-I) on real data is $74.97$ ($80.84$), while current models achieve values of around $30-36$ ($40-50$).
	Since the network used to calculate the SOA values is not part of the training loop the models can not easily overfit to this evaluation metric like they can for the R-precision.
	Furthermore, the results of the SOA evaluation confirm the impression that none of the models is able to generate images with multiple distinct objects of a quality similar to real images.

    \textbf{Impact of the Object Pathway}\ \ To get a clearer understanding of how the evaluation metrics might be impacted by the object pathway we calculate our scores for a different number of generated objects.
	More specifically, we only apply the object pathway for a maximum given number of objects (0, 1, 3, or 10) per image.
	Intuitively, we would assume that without the application of the object pathway the IS and FID should be decreased, since the object pathway is not used to generate any object features and the images should, therefore, consist mostly of background.
	Additionally, we can get an intuition of how important the object pathway is for the overall performance of the network by looking at how it affects the R-precision, SOA-C, and SOA-I.
	
	As \autoref{tab:coco:results} shows, all models perform markedly worse when the object pathway is not used (0 obj).
	We find that the models trained with up to ten objects per image seem to rely more heavily on the object pathway than models trained with three objects per image.
	For models trained with only three objects per image (\textit{OPv2} and \textit{OPv2} + \textit{BBL}) the IS decreases by around $1-2$, the R-precision decreases by around $4-5$, the SOA-C (SOA-I) decreases by around $7-9$ ($11-14$), CIDEr decreases by around $6-8\%$, and the FID increases by around $4-7$.
	On the other hand, models trained with up to 10 objects suffer much more when the object pathway is removed, with the IS decreasing by around $3-6$, the R-precision decreasing by around $9-15$, the SOA-C (SOA-I) decreasing by around $12-18$ ($17-28$), CIDEr decreasing by around $16-30\%$, and the FID increasing by around $10-20$.
	These results indicate that the object pathways are an important part of the model and are responsible for at least some of the improvements compared to the baseline architecture.
	
	\textbf{Impact of Bounding Box Loss}\ \ Adding the bounding box loss to the object pathways has a small negative effect on all scores, but does slightly improve the IoU scores (see \autoref{tab:coco:soa}).
	Note that the weighting of the bounding box loss in the overall loss term was not optimized but simply weighted with the same strength as the matching aware discriminator loss $\mathcal{L}_{D}^{\text{cls}}$.
	It is possible that the positive effect of the bounding box loss could be increased by weighting it differently.
	
	\textbf{Impact of Training on Many Objects}\ \ Training the model with up to ten objects per image has only minor effects on the IS and SOA scores, but improves the FID and R-precision.
	However, we observe that the models trained with only three objects per image slightly decrease in their performance once the object pathway is applied multiple times.
	Usually, the models trained on only three objects achieve their best performance when applying the object pathway three times as at training time.
	Once the model is trained on up to ten objects though, we do not observe this behavior anymore and instead achieve comparable or even better results when applying the object pathway ten times per image.
	
	\textbf{SOA Scores}\ \ \autoref{tab:coco:soa} shows the results for the SOA and SOA-IoU.
	The SOA-I values are consistently higher than the SOA-C values.
	Since the SOA-I is calculated on image average (instead of class average like the SOA-C) it is skewed by objects that often occur in captions and images (e.g. persons, cats, dogs, etc.).
	The SOA values for the most and least common 40 objects show that the models perform much better on the more common objects.
	Actually, most models perform about two times better on the common objects showing their problem in generating objects that are not often observed during training.
	For a detailed overview of how each model performed on the individual labels please refer to the supplementary material.
	
	\begin{figure*}[!t]
	\centering
	\begin{minipage}[c]{.19\linewidth} \centering
		\includegraphics[width=\textwidth]{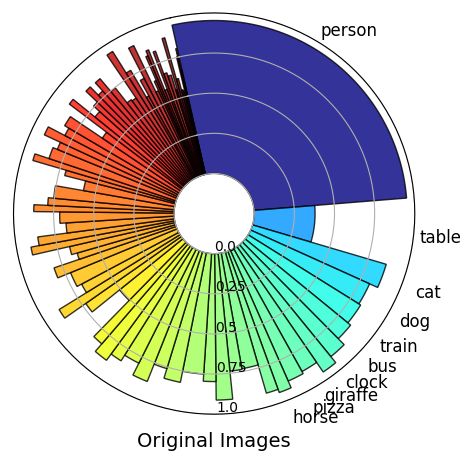}
	\end{minipage}
	\begin{minipage}[t]{.19\linewidth} \centering
		\includegraphics[width=\textwidth]{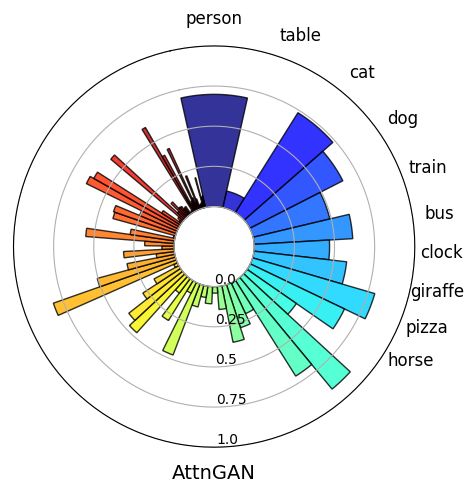}
		\includegraphics[width=\textwidth]{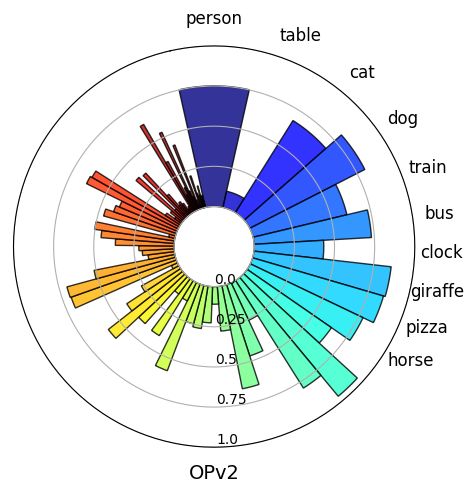}
	\end{minipage}
	\begin{minipage}[t]{.19\linewidth} \centering
		\includegraphics[width=\textwidth]{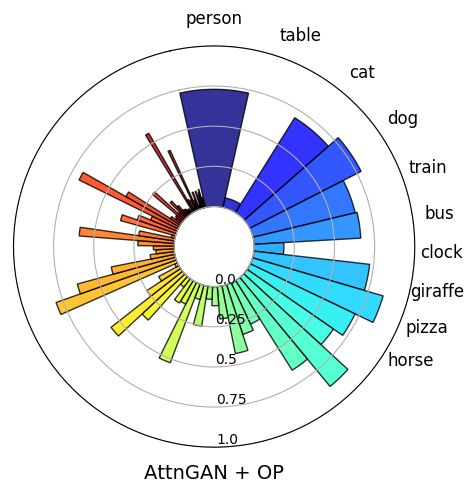}
		\includegraphics[width=\textwidth]{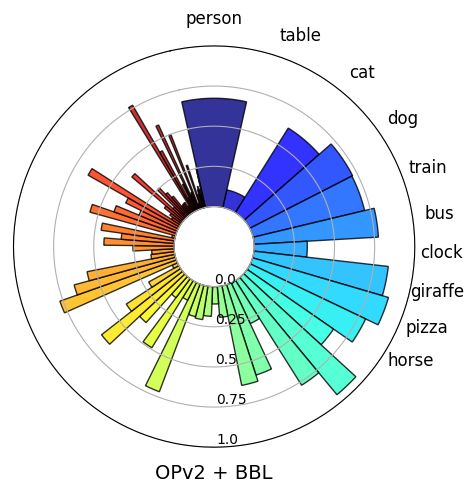}
	\end{minipage}
	\begin{minipage}[t]{.19\linewidth} \centering
		\includegraphics[width=\textwidth]{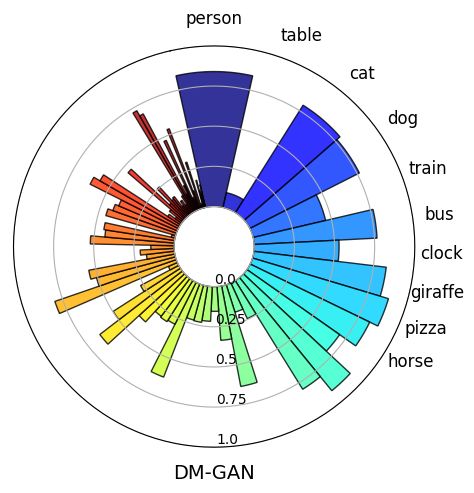}
		\includegraphics[width=\textwidth]{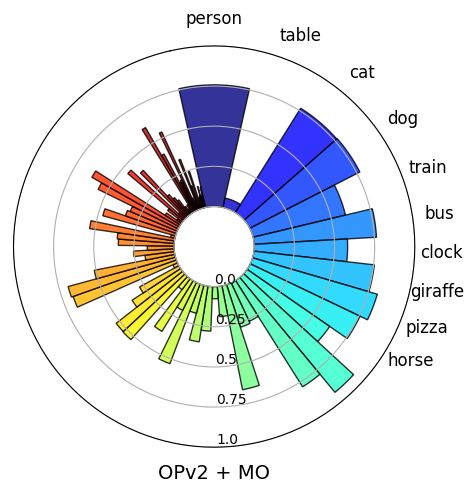}
	\end{minipage}
	\begin{minipage}[t]{.19\linewidth} \centering
		\includegraphics[width=\textwidth]{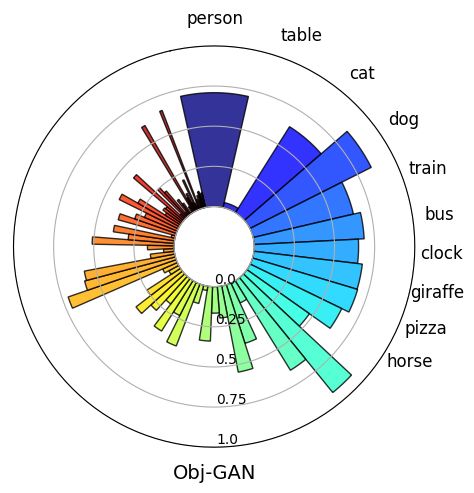}
		\includegraphics[width=\textwidth]{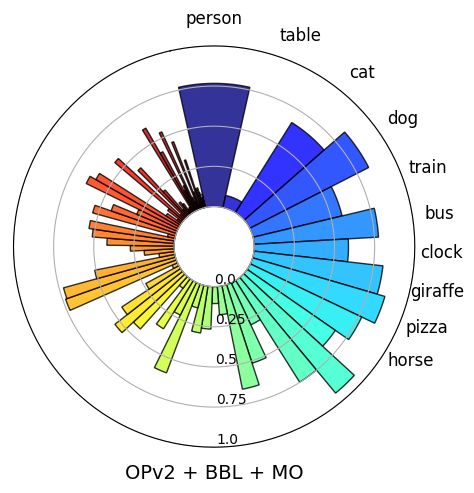}
	\end{minipage}
	\caption{Comparison of SOA scores: SOA per class with degree of a bin reflecting relative frequency of that class.}
	\label{fig:comparison-our-models:soa}
	\end{figure*}
	
	\begin{figure}[!t]
		\centering
		\includegraphics[width=0.48\textwidth]{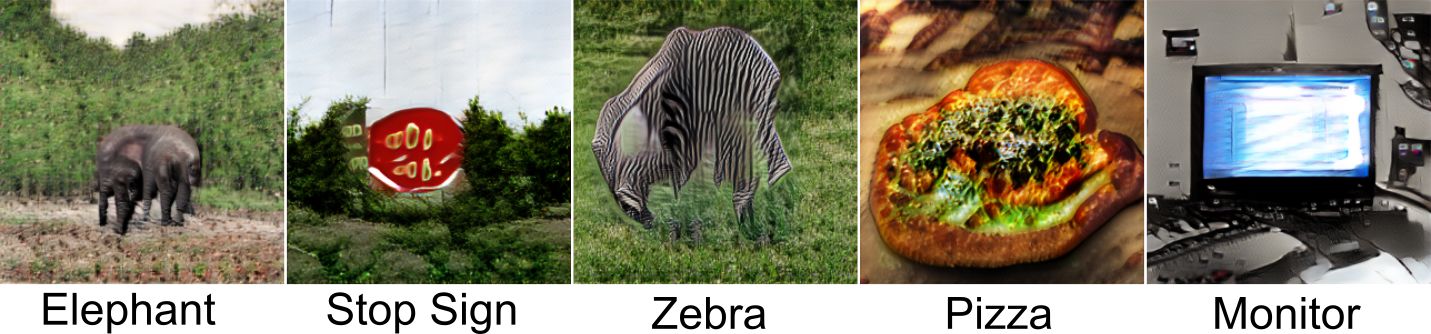}
		\caption{Generated images and objects recognized by the pre-trained object detector (YOLOv3) which was used to calculate the SOA scores. The results highlight that, like most other CNN based object detectors, YOLOv3 focuses much more on texture and less on actual shapes.}
		\label{fig:metric:SOA}
	\end{figure}
	
	When we look at the IoU scores we see that the Obj-GAN \cite{li2019object} achieves by far the best IoU scores (around $0.5$), albeit at the cost of lower SOA scores.
	Our models usually achieve an IoU of around $0.2-0.3$ on average.
	Training with up to ten objects per image and using the bounding box loss slightly increases the IoU.
	However, similar to previous work \cite{hinz2019generating, li2019object} we find that the AttnGAN architecture tends to place salient object features at many locations of the image which affects the IoU scores negatively.

	\begin{figure*}[!t]
	\centering
	\includegraphics[width=\textwidth]{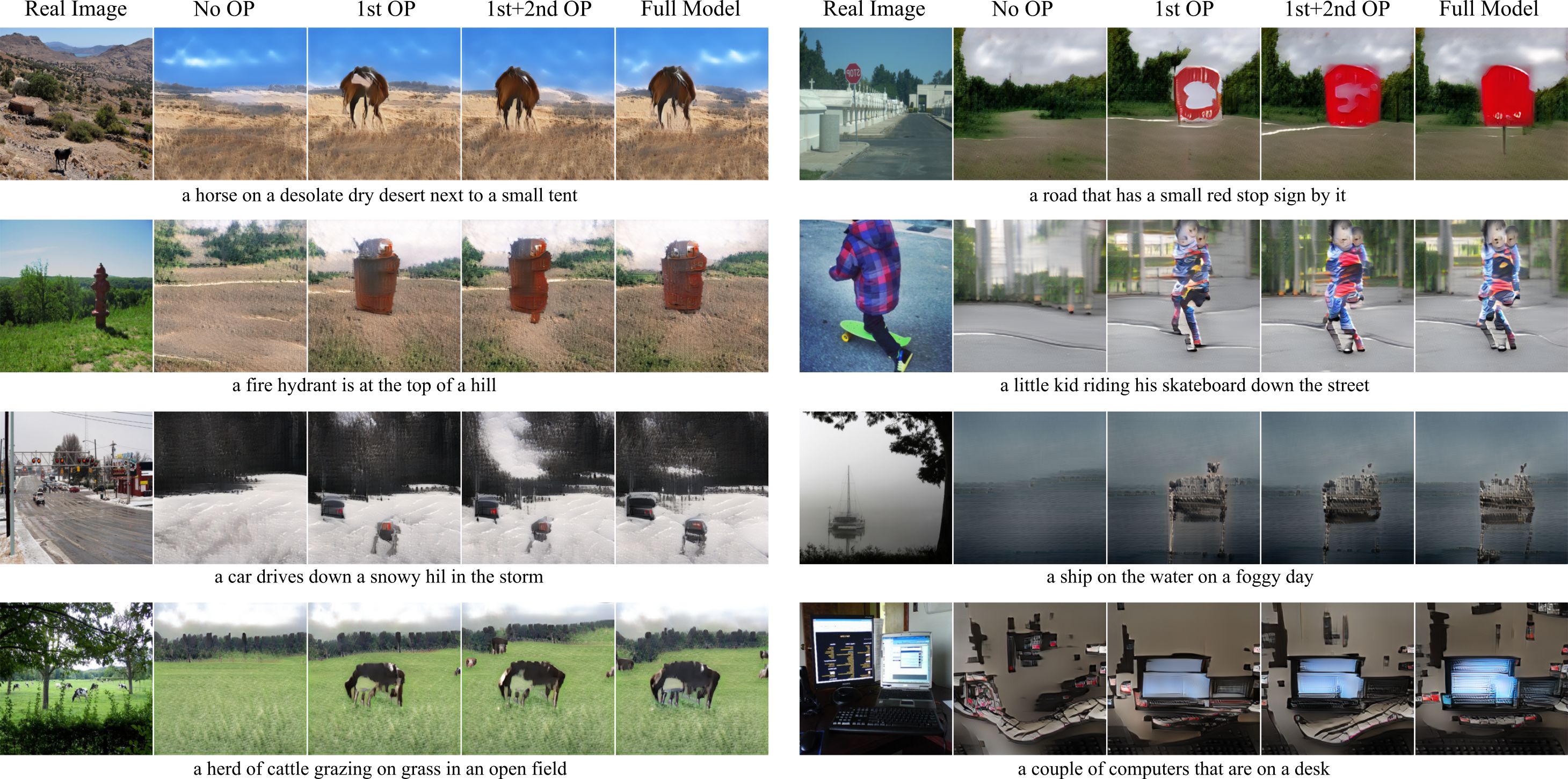}
	\caption{Comparison of images generated by our model (\textit{OP-GAN}) with OPs switched on and off.}
	\label{fig:comparison-op-model}
	\end{figure*}
	
	When looking at the SOA for individual objects (see \autoref{fig:comparison-our-models:soa}) we find that there are objects for which we can achieve very high SOA values (e.g. person, cat, dog, zebra, pizza, etc.).
	Interestingly, we find that all tested methods perform ``good'' or ``bad'' at the same objects.
	For example, all models perform reasonably well on objects such as \textit{person} and \textit{pizza} (many examples in the training set) as well as e.g. \textit{plane} and \textit{traffic light} (few examples in the training set).
	Conversely, all models fail on objects such as \textit{table} and \textit{skateboard} (many examples in the training set) as well as e.g. \textit{hair drier} and \textit{toaster} (few examples in the training set).
	
	We found that objects need to have three characteristics to achieve a high SOA and the highest SOA scores are achieved when objects possess all three characteristics.
	The first important characteristic is easily predictable: the higher the occurrence of an object in the training data, the better (on average) the final performance on this object.
	Secondly, large objects, i.e. objects that usually cover a large part of the image (e.g. \textit{bus} or \textit{elephant}), are usually modeled better than objects that are usually small (\textit{spoon} or \textit{baseball glove}).
	The final and more subtle characteristic is the surface texture of an object.
	Objects with highly distinct surface textures (e.g. \textit{zebra}, \textit{giraffe}, \textit{pizza}, etc.) achieve high SOA scores because the object detection network relies on these textures to detect objects.
	However, while the models are able to correctly match the surface texture (e.g. black and white stripes for a zebra) they are still not capable of generating a realistic-looking shape of many objects.
	As a result, many of these objects possess the ``correct'' surface texture but their shape is more a general ``blob'' consisting of the texture and not a distinct form (e.g. a snout and for legs for a zebra).
	See \autoref{fig:metric:SOA} for a visualization of this.
	
	This is one of the weaknesses of the SOA score as it might give the wrong impression that an 80\% object detection rate means in 80\% of the cases the object is recognizable and of real-world quality.
	This is not the case, as the SOA scores are calculated with a pre-trained object detector which might focus more on texture and less on shapes of objects \cite{geirhos2018imagenet}.
	Consequently, the results of the SOA are more aptly interpreted as cases where a model was able to generate features that an independently pre-trained object detector would classify as a given object.
	The overall quality of the metric is, therefore, strongly dependent on the object detector and future improvements in this area might also lead to more meaningful interpretations of the SOA scores.
	
	\autoref{fig:comparison-our-models} shows images generated by our different models.
	All images shown in this paper were generated without ground truth bounding boxes but instead use generated bounding boxes \cite{li2019object}.
	The first column shows the respective image from the data set, while the next four columns show the generated images.
	We can see that all models are capable of generating recognizable foreground objects.
	It is often difficult to find qualitative differences in the images generated by the different models.
	However, we find that the models using the bounding box loss usually improve the generation of rare objects.
	Training with ten objects per image usually leads to a slightly better image quality overall, especially for images that contain many objects.
	
	As we saw in the quantitative evaluation, the object pathway can have a large impact on the image quality.
	\autoref{fig:comparison-op-model} shows what happens when (some of) the object pathways are not used in the full model (\textit{OPv2} + \textit{BBL} + \textit{MO}).
	Again, the first column shows the original image from the data set and the second column shows images generated without the use any of the object pathways.
	The next three columns show generated images when we consecutively use the object pathways, starting with the lowest object pathway and iteratively adding the next object pathway until we reach the full model.
	When no object pathway is used (first column) we clearly see that only background information is generated.
	Once the first object pathway is added we also get foreground objects and their quality gets slightly better by adding the higher-level object pathways.
	
	\textbf{User Study}\ \ In order to further validate our results, we performed a user study on Amazon Mechanical Turk.
	Similar to other approaches \cite{zhang2017stackgan++,yin2019semantics,hong2018inferring} we sampled 5,000 random captions from the COCO validation set.
	For each caption, we generated one image with each of the following models: our OP-GAN, the AttnGAN \cite{xu2017attngan}, the AttnGAN-OP \cite{hinz2019generating}, the Obj-GAN \cite{li2019object}, and the DM-GAN \cite{zhu2019dm}.
	We showed each user a given caption and the respective five images from the models in random order and asked them to choose the image that depicts the given caption best.
	We evaluated each image caption twice, for a total of 10,000 evaluations with the help of 200 participants.
	
	\begin{table}[bt]
	\centering
	\caption{Human evaluation results (ratio of 1st by human ranking) of five models on the MS-COCO data set given a caption.}
	\label{tab:user-study}
	\setlength{\tabcolsep}{15pt}
	\begin{tabular}{l c}
		\toprule
		AttnGAN-OP \cite{hinz2019generating} & $14.65\% \pm 0.35$ \\
		AttnGAN \cite{xu2017attngan} & $16.80\% \pm 0.43$ \\
		Obj-GAN \cite{li2019object} & $20.96\% \pm 0.33$ \\
		DM-GAN \cite{zhu2019dm} & $22.42\% \pm 0.41$ \\
		OP-GAN (ours) & $\boldsymbol{25.17\% \pm 0.43}$ \\
		\bottomrule
	\end{tabular}
	\end{table}

	\autoref{tab:user-study} shows how often each model was chosen as having produced the best image given a caption (variance was estimated by bootstrap \cite{efron1992bootstrap}).
	This evaluation reveals that the human ranking closely reflects the ranking obtained through the SOA and FID scores.
	One notable exception are the two worst performing models (AttnGAN and AttnGAN-OP), which we measure to perform similar according to the SOA and FID scores, but obtain different results in the user study.
	We find that the IS score is not predictive of the performance in the user study.
	The R-precision and CIDEr are somewhat predictive, but predict a different ranking of the top-three performing models.
	Overall, we find that our OP-GAN performs best according to both the SOA scores and the human evaluation.
	As hypothesized in \autoref{sec:eval-metric} we also observe that the FID and SOA scores are the best predictors for a model's performance in a human user evaluation.
	
	\subsection*{Qualitative Results}
\begin{figure*}[!t]
	\centering
	\includegraphics[width=\textwidth]{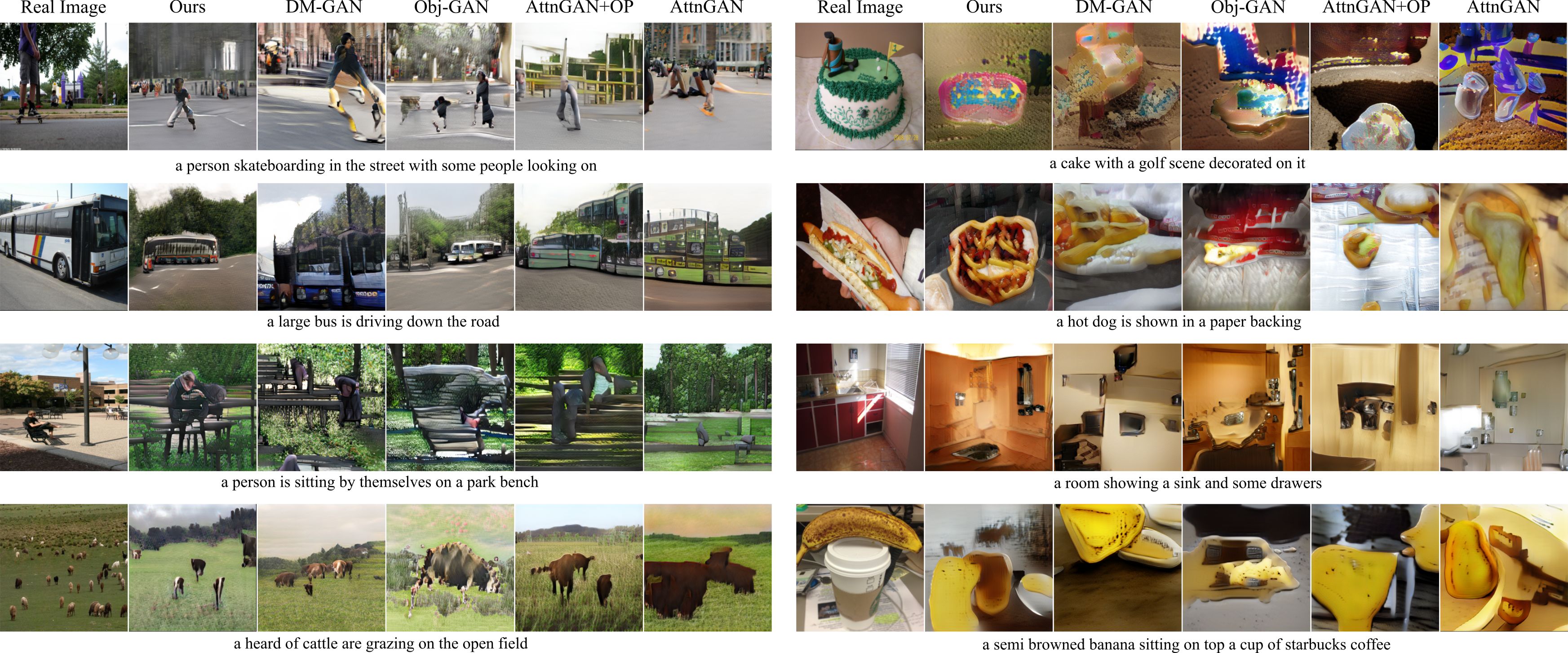}
	\caption{Comparison of images generated by our model (\textit{OP-GAN}) with images generated by other current models.}
	\label{fig:comparison-sota-models}
	\end{figure*}

	\autoref{fig:comparison-sota-models} shows examples of images generated by our model (\textit{OPv2} + \textit{BBL} + \textit{MO}) and those generated by several other models \cite{zhu2019dm, li2019object, hinz2019generating, xu2017attngan}.
	We observe that our model often generates images with foreground objects that are more recognizable than the ones generated by other models.
	For more common objects (e.g. person, bus or plane) all models manage to generate features that resemble the object but in most cases do not generate a coherent representation from these features and instead distribute them throughout the image. 
	As a result, we notice features that are associated with an object but not necessarily form one distinct and coherent appearance of that object.
	Our model, on the other hand, is often able to generate one (or multiple) coherent object(s) from the features, see e.g. the generated images containing a bus, cattle, or the plane.
	
	When generating rare objects (e.g. cake or hot dog) we observe that our model generates a much more distinct object than the other models.
	Indeed, most models fail completely to generate rare objects and instead only generate colors associated with these objects.
	Finally, when we inspect more complex scenes we see that our model is also capable of generating multiple diverse objects within an image.
	As opposed to the other images for ``\textit{room showing a sink and some drawers}'' we can recognize a sink-like shape and drawers in the image generated by our model.
	Similarly, our model can also generate an image containing a reasonable shape of a banana and a cup of coffee, whereas the other models only seem to generate the texture of a banana without the shape and completely ignore the cup of coffee.	
		
\section{Conclusion}
	In this paper, we introduced a novel GAN architecture (\textit{OP-GAN}) that specifically models individual objects based on some textual image description.
	This is achieved by adding object pathways to both the generator and discriminator which learn features for individual objects at different resolutions and scales.
	Our experiments show that this consistently improves the baseline architecture based on quantitative and qualitative evaluations.
	
	We also introduce a novel evaluation metric named \textit{Semantic Object Accuracy} (SOA) which evaluates how well a model can generate individual objects in images.
	This new SOA evaluation allows to evaluate text-to-image synthesis models in more detail and to detect failure and success modes for individual objects and object classes.
	A user study with 200 participants shows that the SOA score is consistent with the ranking obtained by human evaluation, whereas other scores such as the Inceptions Score are not.
	Evaluation of several state-of-the-art approaches using SOA shows that no current approach is able to generate realistic foreground objects for the 80 classes in the COCO data set.
	While some models achieve high accuracy for several of the most common objects, all of them fail when it comes to modeling rare objects or objects that do not have an easily recognizable surface structure.
	However, using the SOA as an evaluation metric on text-to-image models provides more detailed information about how well they perform for different object classes or image captions and is well aligned with human evaluation.
	
\bibliographystyle{IEEEtran}
\bibliography{IEEEabrv,references}

\ifCLASSOPTIONcompsoc
  \section*{Acknowledgments}
\else
  \section*{Acknowledgment}
\fi
	The authors gratefully acknowledge partial support from the German Research Foundation DFG under project CML (TRR 169).
	We also thank the NVIDIA Corporation for their support through the GPU Grant Program.

\ifCLASSOPTIONcaptionsoff
  \newpage
\fi

\begin{IEEEbiographynophoto}{Tobias Hinz}
received his bachelor's degree in Business Informatics from the University of Mannheim in 2014 and his master's degree in Intelligent Adaptive Systems from the University of Hamburg, Germany, in 2016. Since 2017 he is a PhD student at the Knowledge Technology Group at University of Hamburg and since 2019 he is a research associate in the international research centre Crossmodal Learning (TRR-169). His current research focus are generative models, computer vision, and scene understanding. In particular, he is interested in how to learn representations of complex visual scenes in an unsupervised manner.
\end{IEEEbiographynophoto}

\begin{IEEEbiographynophoto}{Stefan Heinrich}
received his Diplom (German MSc) in computer science and cognitive psychology from the University of Paderborn, and his PhD in Computer Science from the Universit\"at Hamburg, Germany. He is a postdoctoral researcher at the International Research Center for Neurointelligence of the University of Tokyo and previously was appointmened as a postdoctoral research associate in the international collaborative research centre Crossmodal Learning (TRR-169). His research interest is located in between artificial intelligence, cognitive psychology, and computational neuroscience. Here, he aims to explore computational principles in the brain, such as timescales, compositionality, and uncertainty, to foster our fundamental understanding of the brain's mechanisms but also to exploit them in developing machine learning methods for intelligent systems.
\end{IEEEbiographynophoto}

\begin{IEEEbiographynophoto}{Stefan Wermter}
is Full Professor at the University of Hamburg, Germany, and Director of the Knowledge Technology Research Group. His main research interests are in the fields of neural networks, hybrid knowledge technology, neuroscience-inspired computing, cognitive robotics, and human-robot interaction. He has been associate editor of the journal `Transactions on Neural Networks and Learning
Systems', is associate editor of `Connection Science' and `International Journal for Hybrid Intelligent Systems', and is on the editorial board of the journals `Cognitive Systems Research', `Cognitive Computation' and `Journal of
Computational Intelligence'. Currently, he serves as co-coordinator of the international collaborative research centre on Crossmodal Learning (TRR-169) and is the coordinator of the European Training Network SECURE on safety for cognitive robots. He is the elected President for the European Neural Network Society for 2020--2022. 
\end{IEEEbiographynophoto}

\appendices
\clearpage
\section*{Information about Captions for SOA}
	\autoref{app:table:label-words} gives a detailed overview of how we chose the captions for each label to calculate the \textit{Semantic Object Accuracy} (SOA) scores.
	The second column shows how many captions we found in total for the given label.
	The third column shows which words we filtered the captions for to obtain captions for the given label.
	This means that we chose all captions that contained at least one of those words as a valid caption for the given label.
	In the fourth column we show (were applicable) which words were explicitly excluded when looking for captions for the given label.
	Finally, the last column shows some examples of ``false positives'', i.e. captions that are included in the set of captions for the given label even though they do not necessarily explicitly ask for the presence of the given label as understood by humans.
	Code to use the SOA can be found here: \url{https://github.com/tohinz/semantic-object-accuracy-for-generative-text-to-image-synthesis}.
	
	\section*{Inspection of YOLO Predictions}
\begin{figure*}[!t]
		\centering
		\includegraphics[width=0.99\textwidth]{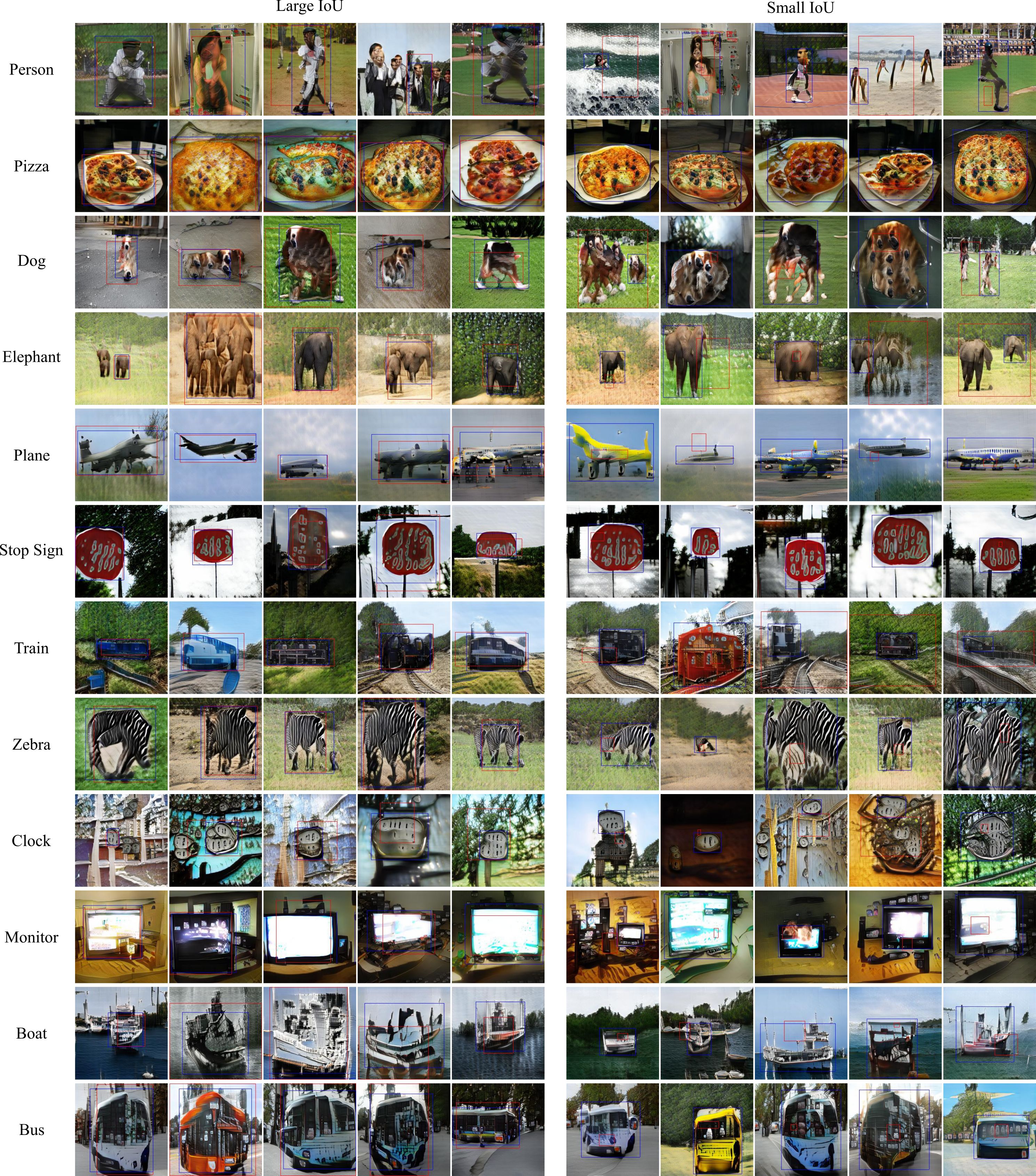}
		\caption{Examples of our model and YOLOv3 predictions on the generated images. The bounding boxes in red are the bounding boxes provided to the network at test time for the given objects. The blue bounding boxes are the bounding boxes provided by YOLOv3 for the given object. When the Intersection over Union (IoU) is small (right column) we observe that this is usually due to the fact that the generated object is much larger than the originally provided bounding box. Only in few cases is the generated object at a completely different location than the provided red bounding box.}
		\label{fig:yolo-eval}
\end{figure*}

	\autoref{fig:yolo-eval} shows generated images with the ground truth bounding boxes (red) provided as input to the model and the bounding boxes detected by YOLO (blue).
	When the Intersection over Union (IoU) is small (right column) we observe that this is usually due to the fact that the generated object is much larger than the originally provided bounding box.
	This agrees with our hypothesis that the reason for the relatively small IoU numbers for our model is because it tends to put salient object features even at locations outside of the provided bounding box.
	Note that our model rarely generates the desired object at a location completely different from the provided bounding box.
	Rather, it tends to increase the object's size, especially when the provided bounding box is small.
	However, we can also see that most objects are not clearly recognizable to humans even though they are ``correctly'' detected by the YOLO network.
	This is in line with our observation that YOLO, like many other CNNs, tend to be focused on textural cues much more than on shapes.
	As a result, future improvements in object detection models can also help increase the information provided by our SOA score.
	
\section*{Model Architecture}
	\autoref{app:tab:architecture:overview} shows our model's architecture. More details and the code can be found here: \url{https://github.com/tohinz/semantic-object-accuracy-for-generative-text-to-image-synthesis}.
	We train our model on four NVIDIA GeForce GTX 1080Ti GPUs.
	Training one model takes between two and four weeks, depending on the exact setting.
	
\section*{Further Results}
	\autoref{app:table:yolov3} and \autoref{app:table:yolov3:ablation} show the detailed results of the YOLOv3 detection network on the individual labels for all models.

\onecolumn
\small
\LTcapwidth=\textwidth
\newcommand{\multiline}[1]{\begin{tabular}{@{}l@{}}#1\end{tabular}}
\begin{longtable}{@{}l c l l l@{}}
\caption{Words that were used to identify given labels in the image caption for the YOLOv3 object detection test (plural of each word also included, different forms of spelling also included).}
\label{app:table:label-words}\\	
		\toprule
		Label & \# Sent. & Words in Captions & Excluded Strings & False Positives \\
		\midrule
		Person & 61586 & \multiline{person, people, human, \\ man, men, woman, \\ women, child, children} & & \multiline{A sign advertising an eatery in \\ which people can eat burgers.} \\
		Dining Table & 7678  & table, desk & & \multiline{A sweet dish is kept in a\\ bowl on a table mat.} \\
		Cat & 6609 & cat, kitten & & \multiline{A double parking meter \\ decorated with cat art}\\
		Dog & 5614 & dog, pup & \multiline{hot dog, hotdog,\\hot-dog, cheese dog,\\chili dog, corn dog} & \multiline{Two stuffed dogs under a blanket \\ looking at a picture book.} \\
		Train & 5397 & train & & \multiline{A red train engine sits\\ on the tracks} \\
		Bus & 4027 & bus & & \multiline{The sign is pointing the\\ direction of the bus route.} \\
		Clock & 3870 & clock & & \multiline{} \\
		Giraffe & 3866 & giraffe & & \multiline{A woman standing in front\\ of a giraffe pen} \\
		Pizza & 3655 & pizza & & \multiline{Would you prefer fresh\\ basil on your pizza or sans basil?} \\
		Horse & 3615 & horse & & \multiline{A close-up of a man\\ hating the horses face.} \\
		Elephant & 3133 & elephant & \multiline{toy elephant,\\stuffed elephant} & \multiline{Outdoor art display of elephant\\ sculptures of various colorings.} \\
		Zebra & 3070 & zebra & & \multiline{An animal that is part horse and\\ part zebra by another horse.} \\
		Bed & 2923 & bed & & \multiline{A large truck has a\\ flat bed trailer attached} \\
		Boat & 2819 & boat, ship & & \multiline{a upside down boat is\\ on top of a big hil} \\
		Toilet & 2796 & toilet & & \multiline{You can pick either toilet\\ stall in this clean restroom.} \\
		Bird & 2691 & bird & & \multiline{a clock with a painting of a\\ bird on a branch on it} \\
		Skateboard & 2665 & skateboard & & \multiline{} \\
		Car & 2650  & car, auto & \multiline{train car, car window,\\side car,  passenger car,\\subway car, car tire,\\ rail car, tram car,\\street car, trolly car} & \multiline{A museum sign showing\\ the main entrance and car park} \\
		Bench & 2633 & bench & &  \\
		Laptop & 2376 & laptop & & \multiline{} \\
		Surfboard & 2270 & surfboard & & \multiline{} \\
		Truck & 2213 & truck & & \multiline{} \\
		Umbrella & 2107 & umbrella & & \multiline{a man playing with a white\\ ball on a red umbrella} \\
		Kite & 2025 & kite & kite board, kiteboard & \multiline{} \\
		Sports Ball & 2001 & ball & & \multiline{Female tennis player looks on as\\she waits for the ball serve} \\
		Cake & 2012 & cake & cupcake & \multiline{} \\
		Cow & 1981 & cow & & \multiline{A young boy sitting on\\top of a cow statue.} \\
		Bicycle & 1920 & bike, bicycle & \multiline{motorbike, motor bike,\\ motorcycle, dirt bike} & \multiline{A man drives his bike taxi\\ with luggage in the back.} \\
		Chair & 1884 & chair & & \multiline{} \\
		Frisbee & 1775 & frisbee & & \multiline{} \\
		Bear & 1740 & bear & \multiline{teddy bear, stuffed bear,\\ care bear, toy bear} & \multiline{a very old panda bear doll\\ with a handkerchief} \\
		Sandwich & 1649 & sandwich & & \multiline{} \\
		Sheep & 1626 & sheep & & \multiline{furniture shaped like sheep\\ on a open field} \\
		Vase & 1597 & vase & & \multiline{} \\
		Bowl & 1570 & bowl & toilet bowl & \multiline{} \\
		Sink & 1529 & sink & & \multiline{} \\
		Stop Sign & 1491 & stop sign & & \multiline{That sign almost looks like\\a stop sign with no words on it.} \\
		Banana & 1466 & banana & & \multiline{} \\
		Monitor & 1437 & monitor, tv, screen & & \multiline{Four cell phone on a wooden table\\ with their screens on.} \\
		Skis & 1419 & skis & & \multiline{} \\
		Hot Dog & 1717 & \multiline{hot dog, chili dog,\\cheese dog, corn dog} & & \multiline{} \\
		Fire Hydrant & 1408 & hydrant & & \multiline{} \\
		Sofa & 1404 & sofa, couch & & \multiline{} \\
		Teddybear & 1284 & teddybear & & \multiline{} \\
		Aeroplane & 1195 & \multiline{plane, jet,\\aircraft} & & \multiline{Mountaineous view as seen\\ from a jet airliner} \\
		Tie & 1062 & tie & to tie & \multiline{} \\
		Tennis Racket & 993 & racket & & \multiline{} \\
		Cell Phone & 956 & cell phone, mobile phone & & \multiline{} \\
		Refrigerator & 949 & refrigerator, fridge & & \multiline{} \\
		Cup & 902 & cup & & \multiline{A table with measuring cups\\and bowls on it} \\
		Broccoli & 840 & broccoli & & \multiline{} \\
		Donut & 805 & donut & & \multiline{} \\
		Bottle & 766 & bottle & & \multiline{A toy hot dog and\\ketchup bottle on a table} \\
		Suitcase & 736 & suitcase & & \multiline{} \\
		Snowboard & 732 & snowboard & & \multiline{} \\
		Book & 731 & book & & \multiline{A large open room has an\\overhead book shelf} \\
		Remote & 670 & remote & & \multiline{} \\
		Traffic Light & 645 & traffic light & & \multiline{} \\
		Keyboard & 603 & keyboard & & \multiline{} \\
		Apple & 510 & apple & pineapple & \multiline{} \\
		Oven & 506 & oven & microwave oven & \multiline{} \\
		Motorcycle & 495 & \multiline{motorcycle, dirt bike,\\motorbike, scooter} & & \multiline{A group of dirt bike racers in a row} \\
		Carrot & 463 & carrot & & \multiline{} \\
		Scissor & 450 & scissors & & \multiline{} \\
		Parking Meter & 430 & parking meter & & \multiline{} \\
		Microwave & 416 & microwave & & \multiline{} \\
		Orange & 378 & oranges & & \multiline{} \\
		Knife & 376 & knife & & \multiline{} \\
		Fork & 363 & fork & & \multiline{A large fork sculpture stands in\\the water as a large boat passes} \\
		Baseball Bat & 322 & baseball bat & & \multiline{} \\
		Toothbrush & 267 & toothbrush & & \multiline{} \\
		Wine Glass & 264 & wine glass & & \multiline{} \\
		Backpack & 220 & backpack, rucksack & & \multiline{} \\
		Spoon & 206 & spoon & & \multiline{} \\
		Handbag & 107 & handbag, purse & & \multiline{Items from a handbag laid\\out neatly on a carpet} \\
		Toaster & 89 & toaster & & \multiline{} \\
		Potted Plant & 81 & potted plant & & \multiline{} \\
		Mouse & 72 & computer mouse & & \multiline{} \\
		Baseball Glove & 39 & baseball glove & & \multiline{} \\
		Hair Drier & 35 & hair drier& & \multiline{} \\
		\bottomrule
\end{longtable}

\begin{table*}[t]
\centering
\setlength{\tabcolsep}{0.2em}
\caption{Overview of the individual layers used in our networks to generate images of resolution $256\times256$ pixels. Values in brackets ($C$, $H$, $W$) represent the tensor's shape. Numbers in the columns after convolutional, residual, or dense layers describe the number of filters / units in that layer. (fs=$x$, s=$y$, p=$z$, BN=$B$) describes the filter size, stride, padding, and batch norm for that convolutional / residual layer. Everything not specifically mentioned or explained (e.g. RNN-Encoder, DAMSM) is the same as in the AttnGAN (Xu, Tao et al., CVPR, 2018).\vspace{-2em}}
\label{app:tab:architecture:overview}
\begin{minipage}[t]{0.45\linewidth}
\begin{tabular}[t]{l | c}
\toprule
Optimizer: Adam & ($\beta_1=0.5$, $\beta_2=0.999$) \\
\multirow{3}{*}{Activation Functions} & Relu ($RL$), Leaky \\
 & RL ($LR$), Gated \\
 & Linear Unit ($GLU$) \\
\textbf{Attention Mask} & see AttnGAN \\

\textbf{Upsample Block} & \\
\ \ Upsampling & Nearest Neighbor \\
\ \ Conv (fs=$3$, s=$1$, p=$1$, BN=$1$) & $X, GLU$ \\

\textbf{Residual Block} & \\
\ \ Conv x 2 (fs=$3$, s=$1$, p=$1$, BN=$1$) & $X, GLU$, $X$ \\
\ \ Add original input to output & \\
\ \ \ \ of previous conv & \\

\textbf{Prepare Label} & \\
\ \ \ \ \ \ Input Shape (Label $\sigma_i$) & $(81,)$ \\
\ \ \ \ \ \ Dense (BN=$1$) & $100, RL$ \\
\ \ \ \ \ \ Reshape & $(100, 1, 1)$ \\
\ \ \ \ \ \ Replicate & $(100, X, X)$ \\

\textbf{Initial Generator} & \\
\ \ \multirow{2}{*}{Global Pathway Input} & noise, sentence \\
\ \ & emb, layout enc \\
\ \ \ \ Input Shape & $(304,)$ \\
\ \ \ \ Dense (BN=$1$) & $49152, GLU$ \\
\ \ \ \ Reshape & $(1536, 4, 4)$ \\
\ \ \ \ Upsample x 2 (fs=$3$, s=$1$, p=$1$) & $768$, $384$ \\
\ \ Object Pathway Input & object labels $\sigma_i$ \\
\ \ \ \ Prepare Label & $(100, 4, 4)$ \\
\ \ \ \ Upsample x 2 (fs=$3$, s=$1$, p=$1$) & $768$, $384$ \\
\ \ \ \ Transform with STN & \\
\ \ Concat Pathways & $(768, 16, 16)$ \\
\ \ Upsample x 2 (fs=$3$, s=$1$, p=$1$) & $192$, $96$ \\
\ \ Output Shape & $(96, 64, 64)$ \\
\textbf{Generator $128\times 128$} & \\
\ \ Global Pathway Input & $(96, 64, 64)$ \\
\ \ \ \ Input Shape & $(96, 64, 64)$ \\
\ \ \ \ Attention Mask & $(96, 64, 64)$ \\
\ \ \ \ Concatenate & $(192, 64, 64)$ \\
\ \ \ \ Residual x 3 & $192$ \\
\ \ \multirow{2}{*}{Object Pathway Input} & object labels $\sigma_i$ \\
\ \  & prev G output \\
\ \ \ \ Input Shape (Label $\sigma_i$) & $(81,), (96, 64, 64)$ \\
\ \ \ \ Prepare Label & $(128, 16, 16)$ \\
\ \ \ \ Extr Obj Feat w/ STN & $(96, 16, 16)$ \\
\ \ \ \ Concatenate & $(224, 16, 16)$ \\
\ \ \ \ Upsample x 2 (fs=$3$, s=$1$, p=$1$) & $192$, $96$ \\
\ \ \ \ Transf Obj Feat w/ STN & $(192, 64, 64)$ \\
\ \ Concat Pathways & $(288, 64, 64)$ \\
\ \ Upsample (fs=$3$, s=$1$, p=$1$) & $96$ \\
\ \ Output Shape & $(96, 128, 128)$ \\

\textbf{Generator $256\times 256$} & \\
\ \ Global Pathway Input & $(96, 128, 128)$ \\
\ \ \ \ Input Shape & $(96, 128, 128)$ \\
\ \ \ \ Attention Mask & $(96, 128, 128)$ \\
\ \ \ \ Concatenate & $(192, 128, 128)$ \\
\ \ \ \ Residual x 3 & $192$ \\
\ \ \multirow{2}{*}{Object Pathway Input} & object labels $\sigma_i$ \\
\ \ & prev G output \\
\ \ \ \ Input Shape (Label $\sigma_i$) & $(81,), (96, 128, 128)$ \\
\ \ \ \ Prepare Label & $(128, 32, 32)$ \\
\ \ \ \ Extr Obj Feat w/ STN & $(96, 32, 32)$ \\
\ \ \ \ Concatenate & $(224, 32, 32)$ \\
\ \ \ \ Upsample x 2 (fs=$3$, s=$1$, p=$1$) & $192$, $96$ \\
\ \ \ \ Transf Obj Feat w/ STN & $(192, 128, 128)$ \\
\ \ Concat Pathways & $(288, 128, 128)$ \\
\ \ Upsample (fs=$3$, s=$1$, p=$1$) & $96$ \\
\ \ Conv (fs=$3$, s=$1$, p=$1$, BN=$1$) & $3, Tanh$ \\
\ \ Output Shape & $(3, 256, 256)$ \\
\bottomrule
\end{tabular}
\end{minipage}
\hspace{0.5cm}
\begin{minipage}[t]{0.45\linewidth}
\begin{tabular}[t]{l | c}
\toprule
Learning Rate & $0.0002$ \\
Training Epochs & $120$ \\
Batch Size & $24$ \\
Z-Dim / Img-Caption-Dim & $100$ / $256$ \\

\textbf{Layout Encoder} & \\
\ \ Input Shape & ($100, 16, 16)$ \\
\ \ Conv (fs=$3$, s=$2$, p=$1$, BN=$0$) & $50, LR$ \\
\ \ Conv x 2 (fs=$3$, s=$2$, p=$1$, BN=$1$) & $25, LR$, $12, LR$ \\
\ \ Output Shape & $(12, 2, 2)$ \\

\textbf{Discriminator $64\times 64$ } &  \\
\ \ Global Pathway & \\
\ \ \ \ Input Shape & $(3, 64, 64)$ \\
\ \ \ \ Conv (fs=$4$, s=$2$, p=$1$, BN=$0$) & $96, LR$ \\
\ \ \ \ Conv (fs=$4$, s=$2$, p=$1$, BN=$1$) & $192, LR$ \\
\ \ \ \ Output Shape & $(192, 16, 16)$ \\
\ \ Object Pathway & \\
\ \ \ \ Input Shape & $(3, 64, 64)$ \\
\ \ \ \ Extract Object Feat w/ STN & $(3, 16, 16)$ \\
\ \ \ \ Concatenate with labels $\sigma_i$ & $(84, 16, 16)$ \\
\ \ \ \ Conv (fs=$4$, s=$1$, p=$1$) & $192, LR$ \\
\ \ \ \ Transform Object Feat w/ STN & $(192, 16, 16)$ \\
\ \ \ \ Output Shape & $(192, 16, 16)$ \\
\ \ Concat Pathways & $(384, 16, 16)$ \\
\ \ Conv x 2 (fs=$4$, s=$2$, p=$1$, BN=$1$) & $384, LR$, $768, LR$ \\
\ \ Concat w/ Sentence Embedding & $(1024, 4, 4)$ \\
\ \ Conv (fs=$3$, s=$1$, p=$1$, BN=$1$) & $768, LR$ \\
\ \ Conv (fs=$4$, s=$4$, p=$1$, BN=$1$) & $1, Sigmoid$ \\

\textbf{Discriminator $128\times 128$ } &  \\
\ \ Global Pathway & \\
\ \ \ \ Input Shape & $(3, 128, 128)$ \\
\ \ \ \ Conv (fs=$4$, s=$2$, p=$1$, BN=$0$) & $96, LR$ \\
\ \ \ \ Conv (fs=$4$, s=$2$, p=$1$, BN=$1$) & $192, LR$ \\
\ \ \ \ Output Shape & $(192, 32, 32)$ \\
\ \ Object Pathway & \\
\ \ \ \ Input Shape & $(3, 128, 128)$ \\
\ \ \ \ Extract Object Feat w/ STN & $(3, 32, 32)$ \\
\ \ \ \ Concatenate with labels $\sigma_i$ & $(84, 32, 32)$ \\
\ \ \ \ Conv (fs=$4$, s=$1$, p=$1$) & $192, LR$ \\
\ \ \ \ Transform Object Feat w/ STN & $(192, 32, 32)$ \\
\ \ \ \ Output Shape & $(192, 32, 32)$ \\
\ \ Concat Pathways & $(384, 32, 32)$ \\
\ \ \multirow{2}{*}{Conv x 4 (fs=$4$, s=$2$, p=$1$, BN=$1$)} & $384, LR$, $768, LR$ \\
 & $1536, LR$, $768, LR$ \\
\ \ Concat w/ Sentence Embedding & $(1024, 4, 4)$ \\
\ \ Conv (fs=$3$, s=$1$, p=$1$, BN=$1$) & $768, LR$ \\
\ \ Conv (fs=$4$, s=$4$, p=$1$, BN=$1$) & $1, Sigmoid$ \\

\textbf{Discriminator $256\times 256$ } &  \\
\ \ Global Pathway & \\
\ \ \ \ Input Shape & $(3, 256, 256)$ \\
\ \ \ \ Conv (fs=$4$, s=$2$, p=$1$, BN=$0$) & $96, LR$ \\
\ \ \ \ Conv (fs=$4$, s=$2$, p=$1$, BN=$1$) & $192, LR$ \\
\ \ \ \ Output Shape & $(192, 64, 64)$ \\
\ \ Object Pathway & \\
\ \ \ \ Input Shape & $(3, 256, 256)$ \\
\ \ \ \ Extract Object Feat w/ STN & $(3, 64, 64)$ \\
\ \ \ \ Concatenate with labels $\sigma_i$ & $(84, 64, 64)$ \\
\ \ \ \ Conv (fs=$4$, s=$1$, p=$1$) & $192, LR$ \\
\ \ \ \ Transform Object Feat w/ STN & $(192, 64, 64)$ \\
\ \ \ \ Output Shape & $(192, 64, 64)$ \\
\ \ Concat Pathways & $(384, 64, 64)$ \\
\ \ \multirow{3}{*}{Conv x 6 (fs=$4$, s=$2$, p=$1$, BN=$1$)} & $384, LR$, $768, LR$ \\
 & $1536, LR$, $3072, LR$ \\
 & $1536, LR$ $768, LR$ \\
\ \ Concat w/ Sentence Embedding & $(1024, 4, 4)$ \\
\ \ Conv (fs=$3$, s=$1$, p=$1$, BN=$1$) & $768, LR$ \\
\ \ Conv (fs=$4$, s=$4$, p=$1$, BN=$1$) & $1, Sigmoid$ \\

\bottomrule
\end{tabular}
\end{minipage}
\end{table*}
\clearpage

\setlength{\tabcolsep}{0.4em}
\begin{longtable}{@{}l | c | c | c | c | c | c | c | c | c | c}
\caption{Results of YOLOv3 detections on generated and original images. Recall provides the fraction of images in which YOLOv3 detected the given object. \emph{IoU} (Intersection over Union) measures the maximum IoU per image in which the given object was detected. No ground truth information besides the caption was used for all measurements.}
\label{app:table:yolov3}\\
\toprule
\multirow{2}{*}{Label} & \multicolumn{2}{c |}{Orig. Img.} & AttnGAN & \multicolumn{2}{c |}{AttnGAN + OP} & DM-GAN & \multicolumn{2}{c |}{Obj-GAN} & \multicolumn{2}{c}{OP-GAN (Ours)} \\
 & Recall & IoU & Recall & Recall & IoU & Recall & Recall & IoU & Recall & IoU \\
\midrule
	Person & $0.953$ & $0.624$ & $0.698$ & $0.730$ & $0.357$ & $0.840$ & $0.708$ & $0.640$ & $\boldsymbol{0.793}$ & $0.289$ \\
	\rowcolor{Gray}Dining Table & $0.379$ & $0.566$ & $0.104$ & $0.061$ & $0.453$ & $0.094$ & $0.031$ & $0.600$ & $\boldsymbol{0.157}$ & $0.495$ \\
	Cat & $0.868$ & $0.644$ & $0.734$ & $0.697$ & $0.264$ & $\boldsymbol{0.790}$ & $0.632$ & $0.653$ & $0.656$ & $0.339$ \\
	\rowcolor{Gray}Dog & $0.813$ & $0.610$ & $0.651$ & $0.778$ & $0.323$ & $0.764$ & $0.846$ & $0.695$ & $\boldsymbol{0.850}$ & $0.355$ \\
	Train & $0.826$ & $0.627$ & $0.491$ & $\boldsymbol{0.654}$ & $0.370$ & $0.463$ & $0.641$ & $0.670$ & $0.561$ & $0.377$ \\
	\rowcolor{Gray}Bus & $0.848$ & $0.651$ & $0.615$ & $0.665$ & $0.511$ & $0.766$ & $0.685$ & $0.804$ & $\boldsymbol{0.793}$ & $0.366$ \\
	Clock & $0.900$ & $0.502$ & $0.469$ & $0.184$ & $0.359$ & $0.528$ & $\boldsymbol{0.649}$ & $0.587$ & $0.587$ & $0.077$ \\
	\rowcolor{Gray}Giraffe & $0.949$ & $0.662$ & $0.581$ & $0.725$ & $0.486$ & $0.829$ & $0.679$ & $0.585$ & $\boldsymbol{0.868}$ & $0.368$ \\
	Pizza & $0.876$ & $0.630$ & $0.793$ & $0.847$ & $0.363$ & $0.883$ & $0.683$ & $0.737$ & $\boldsymbol{0.893}$ & $0.449$ \\
	\rowcolor{Gray}Horse & $0.891$ & $0.611$ & $0.650$ & $0.723$ & $0.528$ & $\boldsymbol{0.827}$ & $0.634$ & $0.685$ & $\boldsymbol{0.827}$ & $0.328$ \\
	Elephant & $0.937$ & $0.647$ & $0.373$ & $0.653$ & $0.522$ & $\boldsymbol{0.705}$ & $0.476$ & $0.737$ & $0.665$ & $0.360$ \\
	\rowcolor{Gray}Zebra & $0.915$ & $0.650$ & $0.902$ & $0.882$ & $0.420$ & $0.909$ & $0.921$ & $0.735$ & $\boldsymbol{0.931}$ & $0.407$ \\
	Bed & $0.732$ & $0.601$ & $0.704$ & $0.661$ & $0.472$ & $\boldsymbol{0.796}$ & $0.655$ & $0.573$ & $0.754$ & $0.444$ \\
	\rowcolor{Gray}Boat & $0.736$ & $0.502$ & $0.211$ & $0.284$ & $0.208$ & $0.244$ & $0.136$ & $0.557$ & $\boldsymbol{0.323}$ & $0.198$ \\
	Toilet & $0.912$ & $0.591$ & $0.281$ & $0.325$ & $0.315$ & $0.178$ & $0.382$ & $0.750$ & $\boldsymbol{0.543}$ & $0.238$ \\
	\rowcolor{Gray}Bird & $0.797$ & $0.551$ & $0.358$ & $0.430$ & $0.284$ & $\boldsymbol{0.637}$ & $0.546$ & $0.612$ & $0.554$ & $0.267$ \\
	Skateboard & $0.822$ & $0.427$ & $0.040$ & $0.119$ & $0.126$ & $0.153$ & $\boldsymbol{0.164}$ & $0.536$ & $0.127$ & $0.116$ \\
	\rowcolor{Gray}Car & $0.752$ & $0.488$ & $0.143$ & $0.202$ & $0.124$ & $\boldsymbol{0.336}$ & $0.196$ & $0.430$ & $0.310$ & $0.102$ \\
	Bench & $0.760$ & $0.547$ & $0.107$ & $0.079$ & $0.311$ & $0.216$ & $\boldsymbol{0.339}$ & $0.637$ & $0.259$ & $0.225$ \\
	\rowcolor{Gray}Laptop & $0.876$ & $0.617$ & $0.071$ & $0.252$ & $0.337$ & $0.229$ & $0.027$ & $0.425$ & $\boldsymbol{0.349}$ & $0.323$ \\
	Surfboard & $0.794$ & $0.414$ & $0.140$ & $0.091$ & $0.218$ & $0.225$ & $0.117$ & $0.548$ & $\boldsymbol{0.321}$ & $0.172$ \\
	\rowcolor{Gray}Truck & $0.835$ & $0.631$ & $0.472$ & $0.524$ & $0.442$ & $0.622$ & $0.413$ & $0.685$ & $\boldsymbol{0.634}$ & $0.341$ \\
	Umbrella & $0.884$ & $0.548$ & $0.074$ & $0.150$ & $0.177$ & $0.292$ & $0.230$ & $0.591$ & $\boldsymbol{0.381}$ & $0.213$ \\
	\rowcolor{Gray}Kite & $0.822$ & $0.410$ & $0.291$ & $0.163$ & $0.310$ & $0.302$ & $0.370$ & $0.384$ & $\boldsymbol{0.414}$ & $0.160$ \\
	Sports Ball & $0.507$ & $0.161$ & $0.112$ & $0.064$ & $0.027$ & $\boldsymbol{0.295}$ & $0.198$ & $0.297$ & $0.165$ & $0.004$ \\
	\rowcolor{Gray}Cake & $0.726$ & $0.570$ & $\boldsymbol{0.471}$ & $0.365$ & $0.206$ & $0.385$ & $0.286$ & $0.626$ & $0.423$ & $0.305$ \\
	Cow & $0.886$ & $0.598$ & $0.425$ & $0.566$ & $0.472$ & $\boldsymbol{0.649}$ & $0.365$ & $0.638$ & $0.614$ & $0.341$ \\
	\rowcolor{Gray}Bicycle & $0.686$ & $0.546$ & $0.281$ & $0.251$ & $0.284$ & $\boldsymbol{0.498}$ & $0.249$ & $0.503$ & $\boldsymbol{0.48}6$ & $0.297$ \\
	Chair & $0.717$ & $0.566$ & $0.175$ & $0.142$ & $0.157$ & $\boldsymbol{0.269}$ & $0.070$ & $0.257$ & $0.258$ & $0.130$ \\
	\rowcolor{Gray}Frisbee & $0.803$ & $0.350$ & $0.025$ & $0.018$ & $0.050$ & $0.061$ & $0.099$ & $0.625$ & $\boldsymbol{0.101}$ & $0.025$ \\
	Bear & $0.638$ & $0.637$ & $\boldsymbol{0.812}$ & $0.794$ & $0.431$ & $0.800$ & $0.712$ & $0.761$ & $0.737$ & $0.341$ \\
	\rowcolor{Gray}Sandwich & $0.674$ & $0.630$ & $0.505$ & $0.634$ & $0.310$ & $0.508$ & $0.585$ & $0.568$ & $\boldsymbol{0.667}$ & $0.402$ \\
	Sheep & $0.910$ & $0.593$ & $0.303$ & $0.403$ & $0.239$ & $0.545$ & $\boldsymbol{0.573}$ & $0.642$ & $0.559$ & $0.251$ \\
	\rowcolor{Gray}Vase & $0.858$ & $0.600$ & $0.114$ & $0.152$ & $0.468$ & $0.175$ & $0.150$ & $0.306$ & $\boldsymbol{0.276}$ & $0.271$ \\
	Bowl & $0.675$ & $0.633$ & $0.315$ & $0.113$ & $0.170$ & $0.212$ & $0.066$ & $0.598$ & $\boldsymbol{0.330}$ & $0.216$ \\
	\rowcolor{Gray}Sink & $0.712$ & $0.431$ & $0.075$ & $0.128$ & $0.127$ & $0.144$ & $0.165$ & $0.340$ & $\boldsymbol{0.184}$ & $0.102$ \\
	Stop Sign & $0.874$ & $0.608$ & $0.183$ & $0.225$ & $0.207$ & $0.522$ & $0.510$ & $0.830$ & $\boldsymbol{0.591}$ & $0.292$ \\
	\rowcolor{Gray}Banana & $0.788$ & $0.578$ & $0.552$ & $\boldsymbol{0.593}$ & $0.208$ & $0.433$ & $0.287$ & $0.572$ & $0.444$ & $0.308$ \\
	Monitor & $0.754$ & $0.594$ & $0.278$ & $0.225$ & $0.477$ & $0.445$ & $0.385$ & $0.759$ & $\boldsymbol{0.606}$ & $0.213$ \\
	\rowcolor{Gray}Skis & $0.576$ & $0.315$ & $0.010$ & $0.023$ & $0.057$ & $0.023$ & $0.023$ & $0.512$ & $\boldsymbol{0.040}$ & $0.146$ \\
	Hot Dog & $0.711$ & $0.621$ & $0.404$ & $0.355$ & $0.227$ & $0.452$ & $0.371$ & $0.671$ & $\boldsymbol{0.592}$ & $0.332$ \\
	\rowcolor{Gray}Fire Hydrant & $0.927$ & $0.613$ & $0.414$ & $0.256$ & $0.388$ & $0.420$ & $0.274$ & $0.666$ & $\boldsymbol{0.426}$ & $0.282$ \\
	Sofa & $0.834$ & $0.584$ & $0.253$ & $0.179$ & $0.259$ & $\boldsymbol{0.397}$ & $0.221$ & $0.470$ & $0.331$ & $0.292$ \\
	\rowcolor{Gray}Teddy Bear & $0.806$ & $0.643$ & $0.637$ & $\boldsymbol{0.688}$ & $0.336$ & $0.615$ & $0.410$ & $0.707$ & $0.455$ & $0.328$ \\
	Aeroplane & $0.916$ & $0.575$ & $0.612$ & $0.382$ & $0.211$ & $0.571$ & $0.297$ & $0.513$ & $\boldsymbol{0.665}$ & $0.318$ \\
	\rowcolor{Gray}Tie & $0.800$ & $0.574$ & $0.138$ & $0.074$ & $\boldsymbol{0.157}$ & $0.095$ & $0.113$ & $0.385$ & $0.117$ & $0.117$ \\
	Tennis Racket & $0.830$ & $0.432$ & $0.019$ & $0.044$ & $0.071$ & $0.048$ & $\boldsymbol{0.141}$ & $0.518$ & $0.058$ & $0.093$ \\
	\rowcolor{Gray}Cell Phone & $0.590$ & $0.513$ & $0.036$ & $0.054$ & $0.067$ & $0.105$ & $0.134$ & $0.563$ & $\boldsymbol{0.264}$ & $0.201$ \\
	Refrigerator & $0.881$ & $0.631$ & $\boldsymbol{0.593}$ & $0.252$ & $0.408$ & $0.456$ & $0.409$ & $0.518$ & $0.558$ & $0.375$ \\
	\rowcolor{Gray}Cup & $0.706$ & $0.586$ & $0.061$ & $0.054$ & $0.022$ & $\boldsymbol{0.131}$ & $0.040$ & $0.430$ & $0.067$ & $0.137$ \\
	Broccoli & $0.756$ & $0.575$ & $0.130$ & $0.137$ & $0.240$ & $0.255$ & $0.248$ & $0.601$ & $\boldsymbol{0.528}$ & $0.267$ \\
	\rowcolor{Gray}Donut & $0.854$ & $0.655$ & $0.076$ & $0.089$ & $0.213$ & $0.138$ & $0.207$ & $0.712$ & $\boldsymbol{0.304}$ & $0.297$ \\
	Bottle & $0.782$ & $0.590$ & $0.072$ & $0.047$ & $0.020$ & $\boldsymbol{0.148}$ & $0.027$ & $0.002$ & $0.069$ & $0.053$ \\
	\rowcolor{Gray}Suitcase & $0.851$ & $0.612$ & $0.049$ & $0.043$ & $0.407$ & $0.070$ & $0.118$ & $0.662$ & $\boldsymbol{0.122}$ & $0.318$ \\
	Snowboard & $0.746$ & $0.411$ & $0.055$ & $0.030$ & $0.085$ & $\boldsymbol{0.101}$ & $0.080$ & $0.356$ & $0.073$ & $0.114$ \\
	\rowcolor{Gray}Book & $0.628$ & $0.500$ & $0.006$ & $0.032$ & $0.340$ & $\boldsymbol{0.064}$ & $0.007$ & $0.390$ & $0.051$ & $0.184$ \\
	Remote & $0.619$ & $0.440$ & $0.014$ & $0.015$ & $0.120$ & $0.123$ & $0.044$ & $0.488$ & $0.038$ & $0.133$ \\
	\rowcolor{Gray}Traffic Light & $0.942$ & $0.450$ & $0.607$ & $0.565$ & $0.409$ & $\boldsymbol{0.724}$ & $0.619$ & $0.559$ & $0.653$ & $0.215$ \\
	Keyboard & $0.783$ & $0.495$ & $0.397$ & $0.083$ & $0.064$ & $\boldsymbol{0.687}$ & $0.095$ & $0.701$ & $0.350$ & $0.154$ \\
	\rowcolor{Gray}Apple & $0.588$ & $0.593$ & $0.054$ & $0.021$ & $0.162$ & $0.119$ & $0.121$ & $0.709$ & $\boldsymbol{0.140}$ & $0.237$ \\
	Oven & $0.699$ & $0.606$ & $0.067$ & $0.074$ & $0.520$ & $0.174$ & $0.055$ & $0.338$ & $\boldsymbol{0.213}$ & $0.304$ \\
	\rowcolor{Gray}Motorcycle & $0.910$ & $0.597$ & $0.422$ & $0.409$ & $0.396$ & $0.476$ & $0.206$ & $0.528$ & $\boldsymbol{0.629}$ & $0.363$ \\
	Carrot & $0.590$ & $0.537$ & $0.081$ & $0.045$ & $0.097$ & $0.083$ & $0.044$ & $0.545$ & $\boldsymbol{0.116}$ & $0.153$ \\
	\rowcolor{Gray}Scissor & $0.654$ & $0.616$ & $0.047$ & $0.066$ & $0.238$ & $0.071$ & $0.024$ & $0.242$ & $\boldsymbol{0.116}$ & $0.261$ \\
	Parking Meter & $0.816$ & $0.600$ & $0.222$ & $0.114$ & $0.323$ & $0.535$ & $\boldsymbol{0.658}$ & $0.759$ & $0.582$ & $0.300$ \\
	\rowcolor{Gray}Microwave & $0.849$ & $0.568$ & $0.066$ & $0.027$ & $0.326$ & $\boldsymbol{0.120}$ & $0.062$ & $0.518$ & $0.074$ & $0.144$ \\
	Orange & $0.826$ & $0.617$ & $0.024$ & $0.113$ & $0.104$ & $0.303$ & $0.084$ & $0.677$ & $\boldsymbol{0.406}$ & $0.208$ \\
	\rowcolor{Gray}Knife & $0.577$ & $0.537$ & $0.017$ & $0.018$ & $0.085$ & $0.015$ & $0.011$ & $0.148$ & $\boldsymbol{0.037}$ & $0.069$ \\
	Fork & $0.675$ & $0.574$ & $0.029$ & $0.083$ & $0.092$ & $0.029$ & $0.052$ & $0.489$ & $\boldsymbol{0.095}$ & $0.124$ \\
	\rowcolor{Gray}Baseball Bat & $0.653$ & $0.397$ & $0.018$ & $0.010$ & $0.022$ & $0.011$ & $\boldsymbol{0.105}$ & $0.395$ & $0.021$ & $0.078$ \\
	Toothbrush & $0.557$ & $0.505$ & $0.036$ & $0.102$ & $0.157$ & $0.025$ & $\boldsymbol{0.107}$ & $0.554$ & $0.091$ & $0.160$ \\
	\rowcolor{Gray}Wine Glass & $0.888$ & $0.583$ & $\boldsymbol{0.194}$ & $0.119$ & $0.086$ & $0.177$ & $0.076$ & $0.275$ & $0.111$ & $0.110$ \\
	Backpack & $0.620$ & $0.529$ & $0.024$ & $0.049$ & $0.000$ & $\boldsymbol{0.107}$ & $0.093$ & $0.000$ & $0.041$ & $0.000$ \\
	\rowcolor{Gray}Spoon & $0.545$ & $0.533$ & $0.069$ & $0.066$ & $0.000$ & $0.055$ & $0.032$ & $0.000$ & $\boldsymbol{0.091}$ & $0.000$ \\
	Handbag & $0.537$ & $0.583$ & $0.014$ & $0.014$ & $0.000$ & $0.042$ & $0.021$ & $0.000$ & $\boldsymbol{0.043}$ & $0.000$ \\
	\rowcolor{Gray}Toaster & $0.093$ & $0.598$ & $0.000$ & $0.000$ & $0.000$ & $\boldsymbol{0.004}$ & $0.000$ & $0.000$ & $0.000$ & $0.000$ \\
	Potted Plant & $0.753$ & $0.574$ & $0.068$ & $0.048$ & $0.000$ & $0.092$ & $0.035$ & $0.000$ & $\boldsymbol{0.112}$ & $0.000$ \\
	\rowcolor{Gray}Mouse & $0.804$ & $0.537$ & $0.076$ & $0.024$ & $0.067$ & $\boldsymbol{0.145}$ & $0.095$ & $0.636$ & $0.096$ & $0.167$ \\
	Baseball Glove & $0.667$ & $0.514$ & $0.006$ & $0.006$ & $0.001$ & $0.042$ & $\boldsymbol{0.083}$ & $0.591$ & $0.020$ & $0.198$ \\
	\rowcolor{Gray}Hair Drier & $0.050$ & $0.158$ & $0.000$ & $0.000$ & $0.000$ & $0.000$ & $0.000$ & $0.000$ & $0.000$ & $0.000$ \\
\bottomrule
\end{longtable}

\vspace{1em}
\setlength{\tabcolsep}{0.4em}
\begin{longtable}{@{}l | c | c | c | c | c | c | c | c}
\caption{Results of YOLOv3 detections on ablations of our model. Recall provides the fraction of images in which YOLOv3 detected the given object. \emph{IoU} (Intersection over Union) measures the maximum IoU per image in which the given object was detected. No ground truth information besides the caption was used for all measurements.}
\label{app:table:yolov3:ablation}\\
\toprule
\multirow{2}{*}{Label} & \multicolumn{2}{c |}{\textit{OPv2}} & \multicolumn{2}{c |}{\textit{OPv2} + \textit{BBL}} & \multicolumn{2}{c |}{\textit{OPv2} + \textit{MO}} & \multicolumn{2}{c}{\textit{OPv2} + \textit{BBL} + \textit{MO}} \\
 & Recall & IoU & Recall & IoU & Recall & IoU & Recall & IoU \\
\midrule
        Person & $0.783$ & $0.279$ & $0.769$ & $0.278$ & $0.789$ & $0.288$ & $0.771$ & $0.286$ \\
        \rowcolor{Gray}Dining Table & $0.095$ & $0.462$ & $0.126$ & $0.453$ & $0.106$ & $0.466$ & $0.106$ & $0.467$ \\
        Cat & $0.699$ & $0.336$ & $0.725$ & $0.330$ & $0.702$ & $0.337$ & $0.697$ & $0.330$ \\
        \rowcolor{Gray}Dog & $0.831$ & $0.342$ & $0.790$ & $0.330$ & $0.745$ & $0.330$ & $0.827$ & $0.351$ \\
        Train & $0.645$ & $0.390$ & $0.699$ & $0.388$ & $0.642$ & $0.389$ & $0.654$ & $0.379$ \\
        \rowcolor{Gray}Bus & $0.756$ & $0.372$ & $0.721$ & $0.372$ & $0.785$ & $0.384$ & $0.802$ & $0.361$ \\
        Clock & $0.489$ & $0.096$ & $0.542$ & $0.130$ & $0.542$ & $0.098$ & $0.401$ & $0.097$ \\
        \rowcolor{Gray}Giraffe & $0.796$ & $0.337$ & $0.853$ & $0.356$ & $0.819$ & $0.353$ & $0.831$ & $0.365$ \\
        Pizza & $0.853$ & $0.428$ & $0.883$ & $0.427$ & $0.837$ & $0.433$ & $0.822$ & $0.437$ \\
        \rowcolor{Gray}Horse & $0.769$ & $0.313$ & $0.774$ & $0.315$ & $0.789$ & $0.331$ & $0.789$ & $0.327$ \\
        Elephant & $0.684$ & $0.368$ & $0.722$ & $0.373$ & $0.658$ & $0.356$ & $0.646$ & $0.363$ \\
        \rowcolor{Gray}Zebra & $0.946$ & $0.393$ & $0.953$ & $0.404$ & $0.955$ & $0.396$ & $0.941$ & $0.406$ \\
        Bed & $0.806$ & $0.457$ & $0.742$ & $0.466$ & $0.747$ & $0.456$ & $0.765$ & $0.464$ \\
        \rowcolor{Gray}Boat & $0.315$ & $0.207$ & $0.224$ & $0.196$ & $0.244$ & $0.232$ & $0.290$ & $0.214$ \\
        Toilet & $0.523$ & $0.252$ & $0.533$ & $0.246$ & $0.455$ & $0.256$ & $0.473$ & $0.250$ \\
        \rowcolor{Gray}Bird & $0.610$ & $0.258$ & $0.650$ & $0.261$ & $0.628$ & $0.249$ & $0.619$ & $0.264$ \\
        Skateboard & $0.162$ & $0.081$ & $0.156$ & $0.076$ & $0.097$ & $0.095$ & $0.131$ & $0.113$ \\
        \rowcolor{Gray}Car & $0.274$ & $0.119$ & $0.236$ & $0.109$ & $0.198$ & $0.129$ & $0.286$ & $0.112$ \\
        Bench & $0.240$ & $0.229$ & $0.180$ & $0.228$ & $0.255$ & $0.236$ & $0.256$ & $0.225$ \\
        \rowcolor{Gray}Laptop & $0.324$ & $0.309$ & $0.237$ & $0.293$ & $0.201$ & $0.299$ & $0.298$ & $0.317$ \\
        Surfboard & $0.268$ & $0.149$ & $0.215$ & $0.152$ & $0.266$ & $0.144$ & $0.266$ & $0.170$ \\
        \rowcolor{Gray}Truck & $0.585$ & $0.341$ & $0.560$ & $0.333$ & $0.590$ & $0.343$ & $0.593$ & $0.338$ \\
        Umbrella & $0.130$ & $0.178$ & $0.189$ & $0.183$ & $0.163$ & $0.187$ & $0.219$ & $0.210$ \\
        \rowcolor{Gray}Kite & $0.354$ & $0.120$ & $0.518$ & $0.123$ & $0.340$ & $0.104$ & $0.427$ & $0.157$ \\
        Cake & $0.448$ & $0.280$ & $0.424$ & $0.295$ & $0.510$ & $0.305$ & $0.486$ & $0.309$ \\
        \rowcolor{Gray}Sports Ball & $0.067$ & $0.005$ & $0.095$ & $0.004$ & $0.192$ & $0.004$ & $0.128$ & $0.004$ \\
        Cow & $0.611$ & $0.298$ & $0.623$ & $0.324$ & $0.621$ & $0.332$ & $0.645$ & $0.340$ \\
        \rowcolor{Gray}Bicycle & $0.401$ & $0.280$ & $0.368$ & $0.245$ & $0.447$ & $0.283$ & $0.472$ & $0.290$ \\
        Chair & $0.138$ & $0.133$ & $0.150$ & $0.134$ & $0.250$ & $0.141$ & $0.262$ & $0.138$ \\
        \rowcolor{Gray}Frisbee & $0.066$ & $0.024$ & $0.052$ & $0.029$ & $0.043$ & $0.035$ & $0.063$ & $0.022$ \\
        Bear & $0.749$ & $0.348$ & $0.754$ & $0.351$ & $0.739$ & $0.345$ & $0.758$ & $0.354$ \\
        \rowcolor{Gray}Sandwich & $0.648$ & $0.380$ & $0.656$ & $0.380$ & $0.613$ & $0.390$ & $0.716$ & $0.394$ \\
        Sheep & $0.617$ & $0.245$ & $0.578$ & $0.217$ & $0.573$ & $0.247$ & $0.596$ & $0.254$ \\
        \rowcolor{Gray}Vase & $0.182$ & $0.181$ & $0.187$ & $0.210$ & $0.154$ & $0.204$ & $0.239$ & $0.220$ \\
        Bowl & $0.238$ & $0.223$ & $0.215$ & $0.202$ & $0.298$ & $0.223$ & $0.300$ & $0.213$ \\
        \rowcolor{Gray}Sink & $0.196$ & $0.131$ & $0.172$ & $0.106$ & $0.206$ & $0.090$ & $0.195$ & $0.113$ \\
        Stop Sign & $0.584$ & $0.279$ & $0.453$ & $0.280$ & $0.494$ & $0.237$ & $0.449$ & $0.270$ \\
        \rowcolor{Gray}Banana & $0.464$ & $0.274$ & $0.517$ & $0.289$ & $0.426$ & $0.280$ & $0.504$ & $0.284$ \\
        Monitor & $0.510$ & $0.209$ & $0.502$ & $0.225$ & $0.535$ & $0.222$ & $0.581$ & $0.225$ \\
        \rowcolor{Gray}Hotdog & $0.481$ & $0.297$ & $0.443$ & $0.305$ & $0.478$ & $0.311$ & $0.576$ & $0.322$ \\
        Skis & $0.021$ & $0.111$ & $0.037$ & $0.121$ & $0.042$ & $0.115$ & $0.039$ & $0.127$ \\
        \rowcolor{Gray}Sofa & $0.274$ & $0.304$ & $0.289$ & $0.284$ & $0.324$ & $0.334$ & $0.270$ & $0.309$ \\
        Fire Hydrant & $0.456$ & $0.290$ & $0.411$ & $0.298$ & $0.386$ & $0.311$ & $0.360$ & $0.295$ \\
        \rowcolor{Gray}Teddy Bear & $0.595$ & $0.339$ & $0.608$ & $0.344$ & $0.582$ & $0.350$ & $0.621$ & $0.341$ \\
        Aeroplane & $0.701$ & $0.315$ & $0.642$ & $0.321$ & $0.599$ & $0.283$ & $0.634$ & $0.310$ \\
        \rowcolor{Gray}Tie & $0.132$ & $0.098$ & $0.085$ & $0.120$ & $0.112$ & $0.103$ & $0.088$ & $0.103$ \\
        Tennis Racket & $0.044$ & $0.073$ & $0.042$ & $0.066$ & $0.070$ & $0.087$ & $0.041$ & $0.095$ \\
        \rowcolor{Gray}Cell Phone & $0.199$ & $0.156$ & $0.100$ & $0.137$ & $0.147$ & $0.150$ & $0.189$ & $0.171$ \\
        Refrigerator & $0.435$ & $0.373$ & $0.438$ & $0.376$ & $0.522$ & $0.366$ & $0.539$ & $0.366$ \\
        \rowcolor{Gray}Cup & $0.047$ & $0.083$ & $0.078$ & $0.111$ & $0.060$ & $0.091$ & $0.078$ & $0.134$ \\
        Broccoli & $0.418$ & $0.243$ & $0.468$ & $0.243$ & $0.448$ & $0.232$ & $0.436$ & $0.261$ \\
        \rowcolor{Gray}Donut & $0.234$ & $0.247$ & $0.248$ & $0.246$ & $0.277$ & $0.287$ & $0.305$ & $0.278$ \\
        Bottle & $0.079$ & $0.023$ & $0.076$ & $0.009$ & $0.103$ & $0.028$ & $0.140$ & $0.026$ \\
        \rowcolor{Gray}Suitcase & $0.100$ & $0.271$ & $0.084$ & $0.289$ & $0.129$ & $0.308$ & $0.117$ & $0.296$ \\
        Book & $0.015$ & $0.141$ & $0.022$ & $0.135$ & $0.033$ & $0.148$ & $0.041$ & $0.142$ \\
        \rowcolor{Gray}Snowboard & $0.085$ & $0.102$ & $0.067$ & $0.105$ & $0.072$ & $0.106$ & $0.089$ & $0.127$ \\
        Remote & $0.060$ & $0.080$ & $0.066$ & $0.096$ & $0.089$ & $0.120$ & $0.051$ & $0.141$ \\
        \rowcolor{Gray}Traffic Light & $0.696$ & $0.160$ & $0.608$ & $0.163$ & $0.703$ & $0.164$ & $0.600$ & $0.175$ \\
        Keyboard & $0.375$ & $0.147$ & $0.494$ & $0.147$ & $0.484$ & $0.167$ & $0.375$ & $0.156$ \\
        \rowcolor{Gray}Oven & $0.141$ & $0.304$ & $0.172$ & $0.308$ & $0.213$ & $0.322$ & $0.206$ & $0.317$ \\
        Apple & $0.227$ & $0.215$ & $0.160$ & $0.163$ & $0.164$ & $0.199$ & $0.160$ & $0.217$ \\
        \rowcolor{Gray}Motorcycle & $0.590$ & $0.376$ & $0.515$ & $0.335$ & $0.420$ & $0.346$ & $0.501$ & $0.345$ \\
        Scissors & $0.045$ & $0.196$ & $0.105$ & $0.264$ & $0.079$ & $0.271$ & $0.119$ & $0.239$ \\
        \rowcolor{Gray}Carrot & $0.080$ & $0.151$ & $0.093$ & $0.163$ & $0.106$ & $0.160$ & $0.106$ & $0.155$ \\
        Parking Meter & $0.553$ & $0.300$ & $0.305$ & $0.263$ & $0.449$ & $0.288$ & $0.481$ & $0.327$ \\
        \rowcolor{Gray}Microwave & $0.150$ & $0.184$ & $0.120$ & $0.190$ & $0.080$ & $0.184$ & $0.078$ & $0.142$ \\
        Orange & $0.323$ & $0.186$ & $0.335$ & $0.179$ & $0.308$ & $0.201$ & $0.314$ & $0.195$ \\
        \rowcolor{Gray}Knife & $0.035$ & $0.090$ & $0.040$ & $0.099$ & $0.028$ & $0.095$ & $0.036$ & $0.053$ \\
        Fork & $0.096$ & $0.084$ & $0.098$ & $0.118$ & $0.067$ & $0.088$ & $0.095$ & $0.121$ \\
        \rowcolor{Gray}Baseball Bat & $0.016$ & $0.056$ & $0.022$ & $0.038$ & $0.019$ & $0.047$ & $0.039$ & $0.050$ \\
        Toothbrush & $0.041$ & $0.170$ & $0.082$ & $0.204$ & $0.063$ & $0.172$ & $0.094$ & $0.181$ \\
        \rowcolor{Gray}Wine Glass & $0.153$ & $0.101$ & $0.160$ & $0.090$ & $0.155$ & $0.080$ & $0.145$ & $0.110$ \\
        Backpack & $0.049$ & $0.000$ & $0.020$ & $0.000$ & $0.045$ & $0.000$ & $0.051$ & $0.000$ \\
        \rowcolor{Gray}Spoon & $0.093$ & $0.000$ & $0.107$ & $0.000$ & $0.075$ & $0.000$ & $0.078$ & $0.000$ \\
        Handbag & $0.001$ & $0.000$ & $0.015$ & $0.007$ & $0.014$ & $0.000$ & $0.026$ & $0.000$ \\
        \rowcolor{Gray}Toaster & $0.006$ & $0.000$ & $0.001$ & $0.000$ & $0.000$ & $0.000$ & $0.001$ & $0.000$ \\
        Mouse & $0.073$ & $0.083$ & $0.077$ & $0.103$ & $0.047$ & $0.108$ & $0.084$ & $0.124$ \\
        \rowcolor{Gray}Potted Plant & $0.065$ & $0.000$ & $0.055$ & $0.000$ & $0.085$ & $0.000$ & $0.075$ & $0.000$ \\
        Baseball Glove & $0.041$ & $0.042$ & $0.029$ & $0.210$ & $0.033$ & $0.211$ & $0.022$ & $0.137$ \\
        \rowcolor{Gray}Hair Drier & $0.002$ & $0.000$ & $0.000$ & $0.000$ & $0.000$ & $0.000$ & $0.002$ & $0.000$ \\
\bottomrule
\end{longtable}

\end{document}